\documentclass[preprint,12pt]{elsarticle}

\usepackage{amsthm}
\usepackage{lineno}
\usepackage{bm} 
\usepackage[colorlinks=true]{hyperref}
\usepackage{wrapfig}
\usepackage{booktabs} 
\usepackage{algorithm}
\usepackage{algpseudocode}
\usepackage{graphicx}
\usepackage{amsmath}
\usepackage{amssymb}
\usepackage{amsfonts}
\usepackage{multirow}
\usepackage{graphicx}
\usepackage{colortbl}
\usepackage{subfigure}
\usepackage{indentfirst}
\usepackage[dvipsnames]{xcolor}
\usepackage{ulem}
\usepackage{url}
\usepackage{verbatim}
\usepackage{adjustbox}
\usepackage{wrapfig}

\graphicspath{{./images/}}

\newtheorem{remark}{Remark}

\journal{arXiv}
\begin{document}

\begin{frontmatter}


\title{Efficient Global-Local Fusion Sampling for Physics-Informed Neural Networks}

\author[first]{Jiaqi Luo}
\ead{jqluo@suda.edu.cn}
\affiliation[first]{organization={School of Mathematical Sciences, Soochow University},
            city={Suzhou},
            postcode={215006}, 
            state={Jiangsu Province},
            country={China}}

\author[second]{Shixin Xu}
\ead{shixin.xu@dukekunshan.edu.cn}
\affiliation[second]{organization={Zu Chongzhi Center , Duke Kunshan University},
            city={Kunshan},
            postcode={215000}, 
            state={Jiangsu Province},
            country={China}}

\author[third,forth]{Zhouwang Yang\corref{cor}}
\cortext[cor]{Corresponding author}
\ead{yangzw@ustc.edu.cn}
\affiliation[third]{organization={University of Science and Technology of China},
            city={Hefei},
            postcode={230026}, 
            state={Anhui Province},
            country={China}}
\affiliation[forth]{organization={Key Laboratory of the Ministry of Education for Mathematical Foundations and Applications of Digital Technology},
            city={Hefei},
            postcode={230026}, 
            state={Anhui Province},
            country={China}}

\begin{abstract}
The accuracy of Physics-Informed Neural Networks (PINNs) critically depends on the placement of collocation points, as the PDE loss is approximated through sampling over the solution domain. Global sampling ensures stability by covering the entire domain but requires many samples and is computationally expensive, whereas local sampling improves efficiency by focusing on high-residual regions but may neglect well-learned areas, reducing robustness. We propose a \textit{Global–Local Fusion (GLF) Sampling Strategy} that combines the strengths of both approaches. Specifically, new collocation points are generated by perturbing training points with Gaussian noise scaled inversely to the residual, thereby concentrating samples in difficult regions while preserving exploration. To further reduce computational overhead, a lightweight linear surrogate is introduced to approximate the global residual-based distribution, achieving similar effectiveness at a fraction of the cost. Together, these components, residual-adaptive sampling and residual-based approximation, preserve the stability of global methods while retaining the efficiency of local refinement. Extensive experiments on benchmark PDEs demonstrate that GLF consistently improves both accuracy and efficiency compared with global and local sampling strategies. This study provides a practical and scalable framework for enhancing the reliability and efficiency of PINNs in solving complex and high-dimensional PDEs.

\end{abstract}




\end{frontmatter}


\section{Introduction}
\label{s:intro}

Solving partial differential equations (PDEs) is essential in many scientific and engineering fields, including fluid dynamics, material science, and heat transfer. Traditional numerical methods \cite{thomas2013numerical, hughes2012finite, shen2011spectral} are often hindered by the curse of dimensionality \cite{hu2024tackling, powell2007approximate} and can be computationally expensive, particularly for high-dimensional problems or complex geometries. In contrast, deep learning-based methods \cite{long2018pde, long2019pde, sirignano2018dgm, weinan2018deep, zang2020weak} are mesh-free and have proven to be efficient approaches for tackling high-dimensional problems. The mesh-free nature allows these models to handle domains with irregular shapes, offering significant improvements in flexibility and computational efficiency. Among these approaches, Physics-Informed Neural Networks (PINNs) \cite{raissi2019physics, karniadakis2021physics} have emerged as a powerful framework for solving PDEs \cite{lu2021deepxde, pang2019fpinns, zhang2019quantifying} by embedding physical laws directly into the neural network’s training process, and they have been successfully applied to diverse problems in computational science and engineering \cite{raissi2020hidden, yazdani2020systems, lu2021physics}. 

Unlike grid-based numerical methods, PINNs approximate PDE solutions by minimizing a loss function that incorporates data, boundary conditions, and PDE residuals. The core idea is that the network is trained to satisfy both observed data and physical constraints, with PDE residuals forming the primary contribution to the loss. Mathematically, the loss in PINNs can be formulated as an integral of the PDE residuals over the solution domain. However, these integrals are generally intractable and must be approximated numerically through discrete sampling. As a result, the placement of collocation points directly determines the quality of this approximation. Consequently, the design of efficient and informative sampling strategies becomes a critical factor influencing both the accuracy and convergence behavior of PINN training.

Current sampling strategies in PINNs can generally be divided into two categories: \textit{Global Sampling} and \textit{Local Sampling}. \textit{Global Sampling} \cite{lu2021deepxde,nabian2021efficient,jiao2024gaussian,gao2023failure} distributes collocation points uniformly or randomly across the solution domain, and then selects points based on their residuals. This approach helps maintain the accuracy and stable convergence of the solution across the entire domain by preventing unsampled regions from being neglected. However, such exhaustive sampling is computationally demanding, as a large number of points are needed to cover the domain, particularly in high-dimensional problems. In addition, random global sampling often wastes points in well-learned regions, which reduces efficiency.
By contrast, \textit{Local Sampling} \cite{luo2025imbalanced,zeng2022adaptive,wang2025non,mao2023physics} concentrates collocation points in areas with large residuals, where the model struggles most. This makes better use of computational resources and accelerates convergence by targeting the difficult regions. Nevertheless, local methods risk overfitting to these regions and may cause fluctuations elsewhere when other parts of the domain are underrepresented.
In summary, global sampling promotes stable solution quality across the whole domain but is computationally expensive, while local sampling improves efficiency but may compromise robustness by neglecting regions where the residuals are small.
Hence, there is a clear need for a sampling strategy that achieves both global stability and local refinement, while maintaining computational efficiency.

In this paper, we propose a \textit{Global-Local Fusion Sampling Strategy, GLF}. 
The method perturbs each training point with scaled Gaussian noise, where the scaling factor is inversely proportional to the residual. 
This ensures that regions with large residuals receive tighter local refinement while regions with small residuals spread more broadly to maintain exploration, thereby inheriting the adaptivity of local sampling.
Moreover, rather than computing a global residual-based probability distribution, which is effective but computationally costly, we introduce a lightweight approximation that assigns the residual of each training point to its neighborhood, achieving a similar effect at substantially lower cost. 
Overall, the strategy integrates the complementary advantages of global and local approaches by reducing reliance on dense global sampling, maintaining stability across the domain, and preserving the efficiency of local refinement.
Extensive numerical experiments on representative PDEs demonstrate that the proposed strategy consistently improves both accuracy and efficiency compared to global or local sampling schemes.

The main contributions are summarized as follows:
\begin{itemize}
   \item We propose a residual-driven global–local fusion (GLF) sampling strategy for Physics-Informed Neural Networks.
   \item The proposed sampling method perturbs training points with Gaussian noise scaled inversely to the residual magnitude, thereby allocating denser samples to high-residual regions while maintaining sufficient exploration in well-learned areas, akin to local refinement strategies.
   \item We introduce a lightweight approximation of the residual distribution that avoids costly residual recalculations yet achieves a comparable effect to global residual-based approaches.
   \item Across representative PDE benchmarks, GLF consistently improves accuracy and sample efficiency over both global and purely local sampling baselines.
\end{itemize}


\section{Preliminaries}
\label{s:pre}
\subsection{Physics-informed neural networks}
Let $\Omega \subseteq \mathbb{R}^d$ be a spatial domain and $\mathbf{x} \in \Omega$ be a spatial variable.
The PDE problem is to find a solution $u(\mathbf{x})$ such that
\begin{equation}
\label{e.pde}
\begin{aligned} 
&\mathcal{L}u(\mathbf{x}) = f(\mathbf{x}), ~\forall{\mathbf{x}} \in \Omega, \\ 
&\mathcal{B}u(\mathbf{x}) = g(\mathbf{x}), ~\forall{\mathbf{x}} \in \partial\Omega,
\end{aligned}  
\end{equation}
where $\mathcal{L}$ is the partial differential operator, $\mathcal{B}$ is the boundary operator, $f(\mathbf{x})$ is the source function, and $g(\mathbf{x})$ represents the boundary conditions.

We consider a general form of a PINN, where a neural network is used to approximate a function that satisfies both the PDE and the observed data.

Let $\hat{u}(\mathbf{x}; \Theta)$ be the neural network approximated solution with parameters $\Theta$. 
The PINN is trained to minimize the following loss function:
\begin{equation}
\label{e.obj}
   \min_{\Theta} \mathbb{E}_{\mathbf{x}\in \Omega}\|\mathcal{L}\hat{u}(\mathbf{x}; \Theta)-f(\mathbf{x})\| + \lambda\ \mathbb{E}_{\mathbf{x}\in \partial\Omega}\|\mathcal{B}\hat{u}(\mathbf{x}; \Theta)-g(\mathbf{x})\|, 
\end{equation}
where $\|\cdot\|$ is usually the $L_2$-norm and $\lambda$ is a parameter for weighting the sum.

The expectations in Eq.\eqref{e.obj} are integrals over $\Omega$ and $\partial \Omega$. Since these integrals are generally intractable, they are approximated by Monte Carlo sampling. Specifically, if we select $N_\Omega$ collocation points ${ \mathbf{x}_i}$, $i=1, \cdots, N_\Omega$ in $\Omega$ and $N_{\partial\Omega}$ boundary points ${ \mathbf{x}_j^\partial}$, $j=1, \cdots, N_{\partial\Omega}$ on $\partial \Omega$, then the empirical loss becomes
\begin{equation}
\label{e.empirical}
\min_{\Theta} \frac{1}{N_\Omega}\sum_{i=1}^{N_\Omega} |\mathcal{L}\hat{u}(\mathbf{x}_i; \Theta)-f(\mathbf{x}_i)|^2
	+ \lambda \frac{1}{N_{\partial\Omega}}\sum_{j=1}^{N_{\partial\Omega}} |\mathcal{B}\hat{u}(\mathbf{x}_j^\partial; \Theta)-g(\mathbf{x}_j^\partial)|^2.
\end{equation}

Eq.\eqref{e.empirical} highlights that sampling plays a central role in PINNs: the choice of collocation points directly determines how well the underlying integrals are approximated. Poorly chosen samples can bias the residual estimation or lead to instability, while oversampling in well-learned regions wastes computational resources. Thus, designing efficient sampling strategies is critical for achieving both accurate and efficient PINN training.

\subsection{Related work}
Global and local sampling methods for PINNs can both be divided into two categories: those that change the number of collocation points and those that keep it fixed.

\paragraph{Global sampling methods}
RAR \cite{lu2021deepxde} was the first work to demonstrate that adaptively adjusting collocation points can improve the training efficiency of PINNs. The authors used a dense set of random points in the entire domain to evaluate residuals and then added the points with the largest residuals into the training set. 
To mitigate overfitting caused by point clustering in RAR, Hanna et al. \cite{hanna2022residual} proposed constructing a residual-based probability density function (PDF) and sampling new points from this PDF.
RAR-D \cite{wu2023comprehensive} extended RAR by repeatedly adding new points to the training set based on a residual distribution computed from a dense set of random points.
DAS-PINNs \cite{tang2023pinns} employed a generative network to approximate the residual PDF and introduced two strategies for adjusting collocation points: replacing all points in the training set or gradually adding new ones.
FI-PINNs \cite{gao2023failure} followed a related idea, defining a failure probability based on residuals and adaptively adding collocation points in failure regions via a failure-informed enrichment technique. 
Similarly, GAS \cite{jiao2024gaussian} introduced a Gaussian mixture model to approximate the residual distribution and then generated more points around high-residual regions. 

Other global methods keep the number of collocation points unchanged. 
For instance, Nabian et al. \cite{nabian2021efficient} proposed an importance sampling method, where each collocation point is linked to its nearest seed, and its loss is set equal to the seed’s loss. Sampling is then performed according to the resulting distribution, without recomputing global residuals and changing the total number of points. 
R3 sampling \cite{daw2022mitigating} selected points whose residuals exceeded a threshold from a uniform set of collocation points, discarding the others. This process repeats without changing the number of collocation points. 
Gao et al. \cite{gao2023active} used active learning: at each iteration, new samples are generated, their acceptance rates computed based on residuals, and points are either replaced or retained accordingly.
Subramanian et al. \cite{subramanian2022adaptive} proposed another adaptive method that constructs a probability distribution from either residuals or their gradients. The algorithm samples part of the collocation points according to this distribution and part uniformly across the domain, thereby balancing local adaptivity with global coverage. 
RAD \cite{wu2023comprehensive} also relied on dense random points to compute a residual distribution but replaced all collocation points in each update step while keeping the total number unchanged.

\paragraph{Local sampling methods}
Zeng et al. \cite{zeng2022adaptive} decomposed the domain into non-overlapping subdomains and increased sampling density in regions with large errors. Mao et al. \cite{mao2023physics} applied a similar strategy by dividing the domain into subdomains and adding new points in regions with high residuals and gradients.

RSmote \cite{luo2025imbalanced} was a number-fixed method that theoretically demonstrates residuals are typically localized. It applied imbalanced learning to allocate more points to difficult regions and fewer to well-learned areas, focusing resources effectively without increasing the total number of points. Nurwas \cite{wang2025non} introduced a non-uniform random walk strategy that concentrates points in complex regions without requiring an explicit probability density function. This approach maintained a fixed number of samples and achieved low computational cost.

\section{Global-Local Fusion Sampling}
\label{s:method}

Our proposed sampling strategy integrates the strengths of global and local approaches. See Fig.~\ref{f.process} for an overview. The method consists of two key components:
\begin{figure}[!ht]
    \centering
    \subfigure[Compute the residuals]{\includegraphics[width=0.45\linewidth]{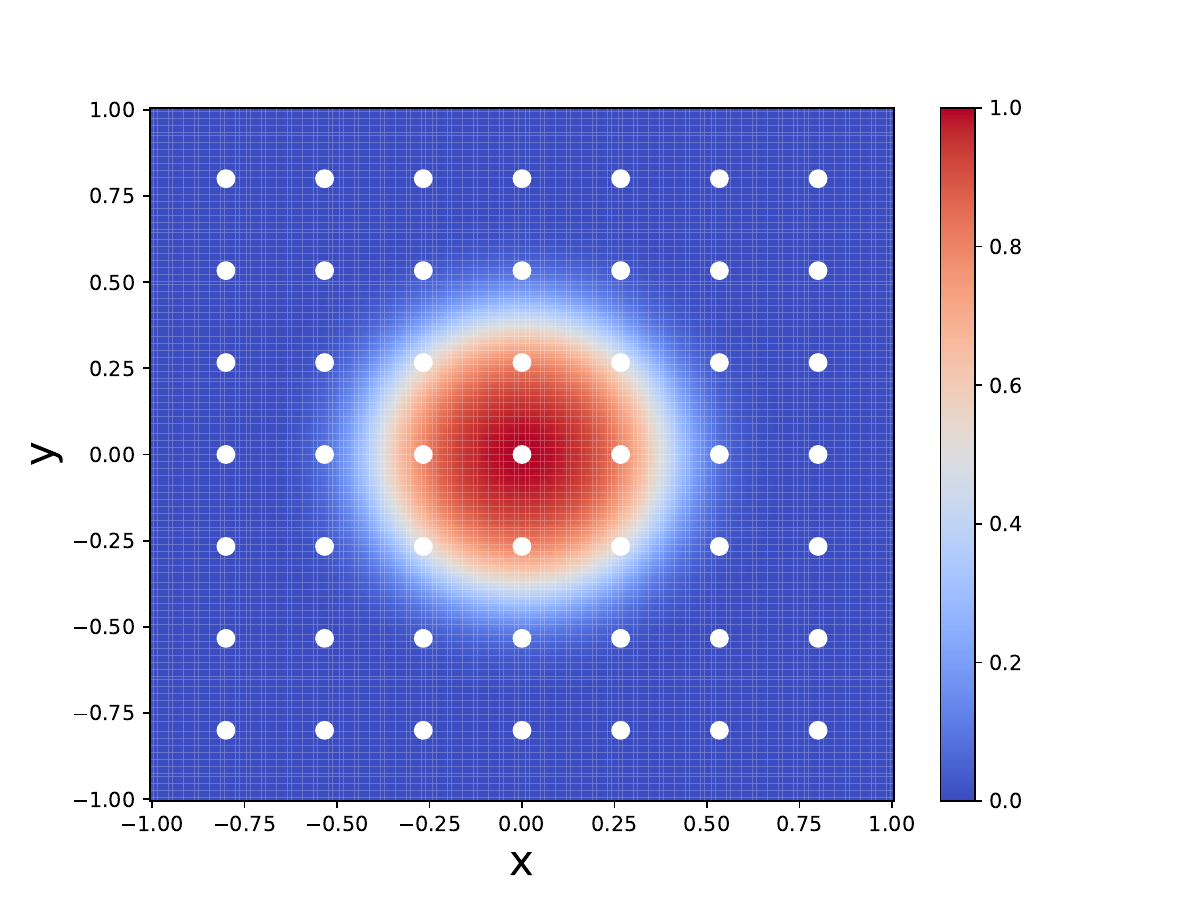}}
    \subfigure[Generate the candidate pool]{\includegraphics[width=0.45\linewidth]{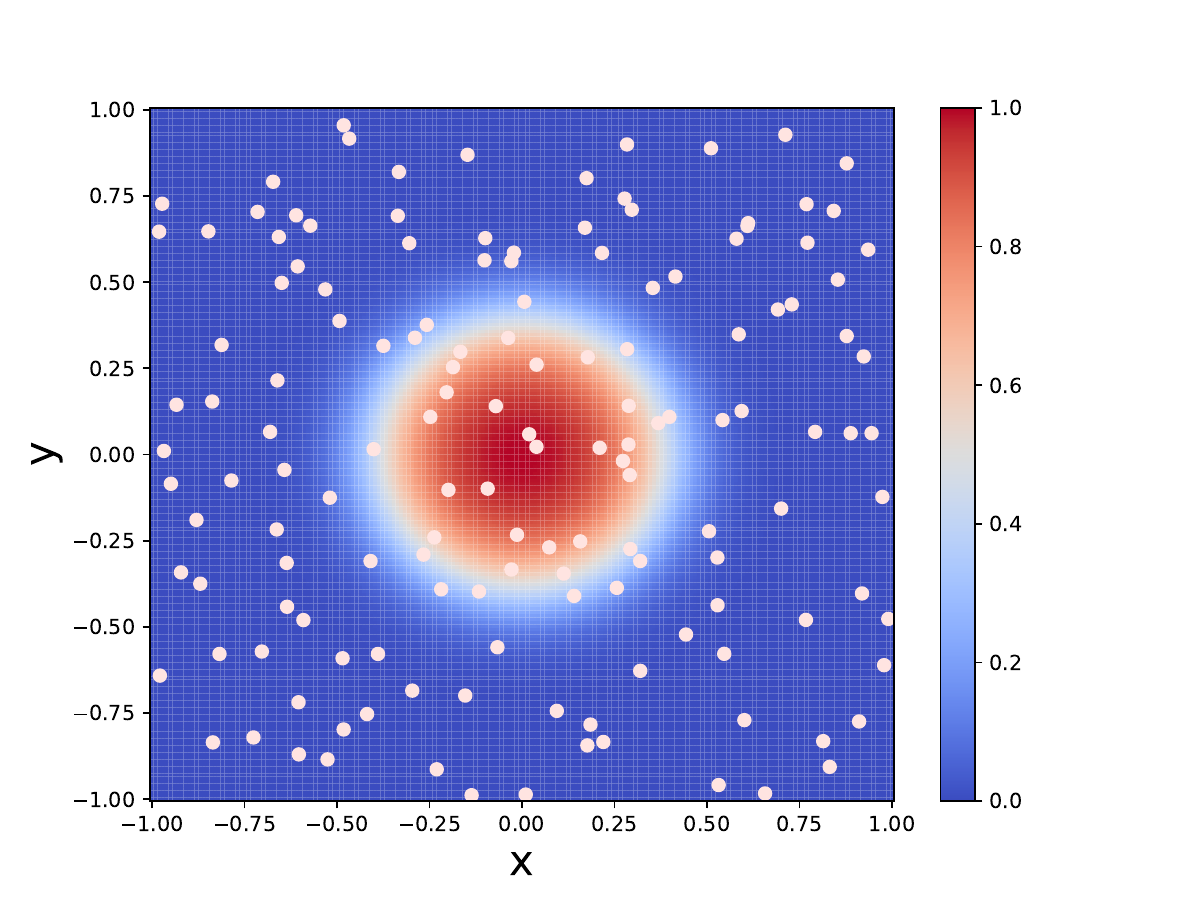}}
    \subfigure[Assign the residual values]{\includegraphics[width=0.45\linewidth]{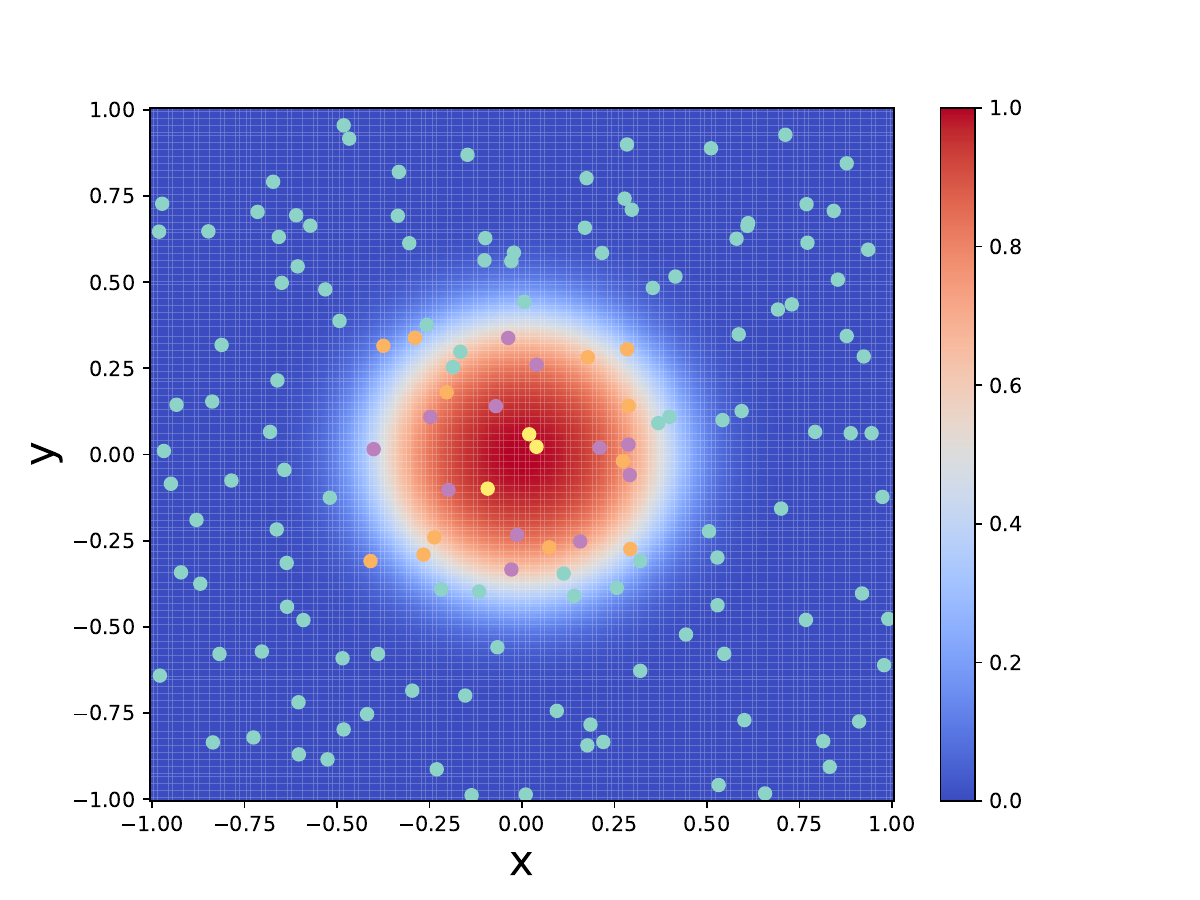}}
    \subfigure[Resample the data]{\includegraphics[width=0.45\linewidth]{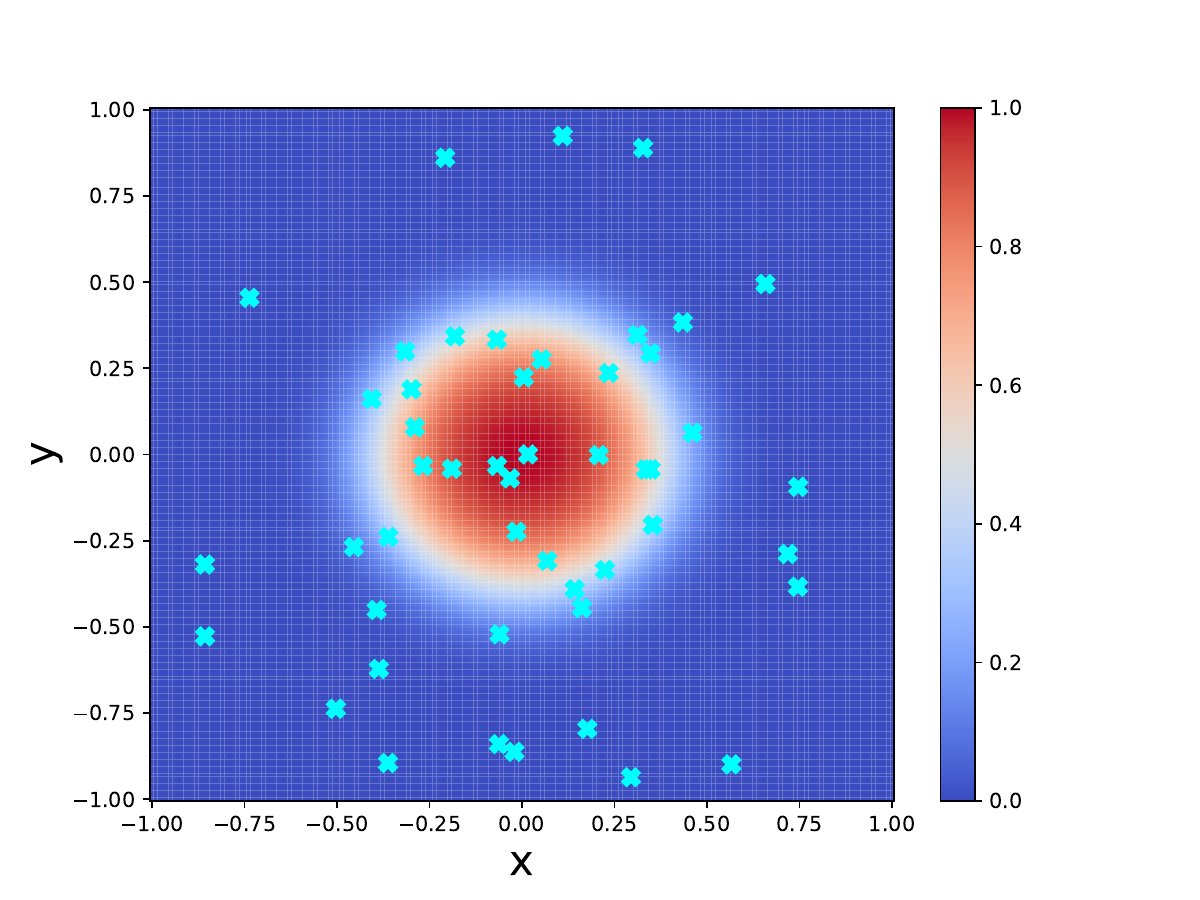}}
    \caption{Algorithm flow. (a) Compute the residuals at training points. (b) Generate $M$ candidate points around each training point (anchor). (c) Assign each candidate the residual value of its anchor point (colors indicate different residual values). (d) Apply the approximate distribution to sample $N$ new points to update the training data.}
    \label{f.process}
\end{figure}

\subsection{Local Residual-driven Probabilistic Sampling}
Let $X = \{\mathbf{x}_i\}_{i=1}^N$ denote the current set of training points. For each point $\mathbf{x}_i$, we compute its residual value 
\begin{equation}
\label{e.residual}
    r(\mathbf{x}_i)=\| \mathcal{L}\hat{u}(\mathbf{x}_i; \Theta) - f(\mathbf{x}_i)\|,
\end{equation}
where the notations follow Eq.\eqref{e.obj}.

A scaling factor $h(\mathbf{x}_i)$ is defined for each training point $\mathbf{x}_i$ (anchor), inversely proportional to its residual:
\begin{equation}
\label{e.radius}
    h(\mathbf{x}_i) = \frac{\alpha}{r(\mathbf{x}_i) + \epsilon},
\end{equation}
where $\alpha$ is a scale control coefficient that adjusts the overall magnitude of $h(\mathbf{x}_i)$ and $\epsilon$ is a small constant for stability. 

\begin{figure}[!h]
    \centering
    \subfigure[Standard normal distribution]{\includegraphics[width=0.32\linewidth]{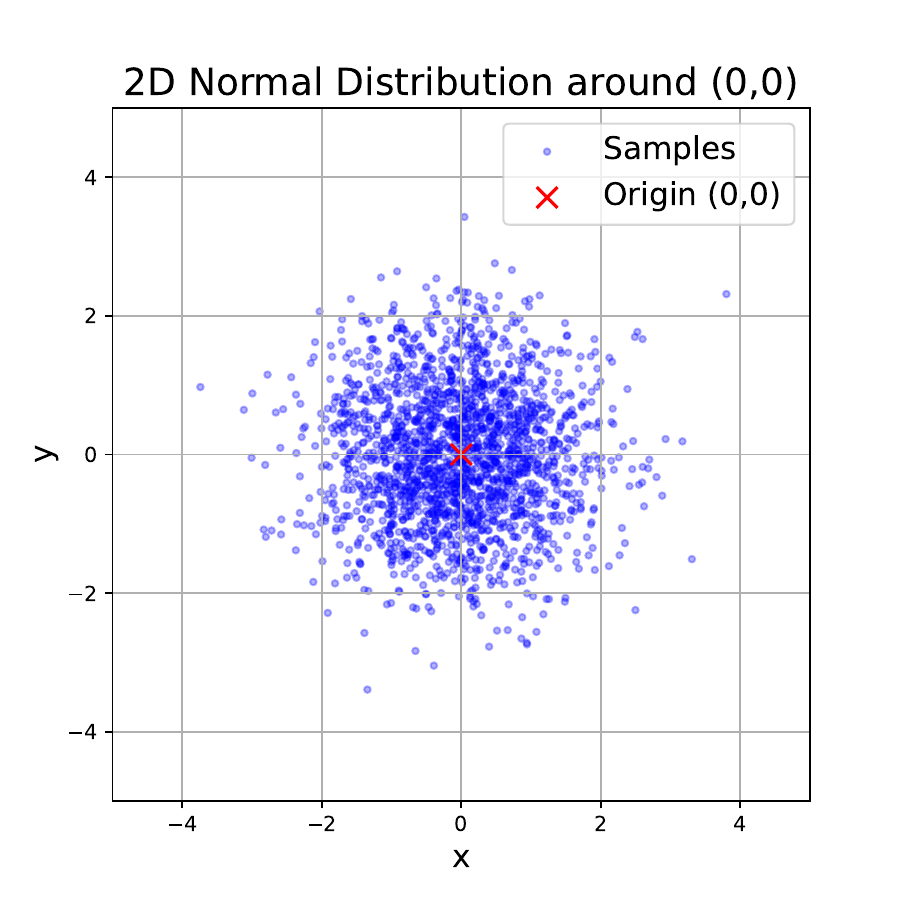}}
    \subfigure[Large scaling factor $h(\mathbf{x})$]{\includegraphics[width=0.32\linewidth]{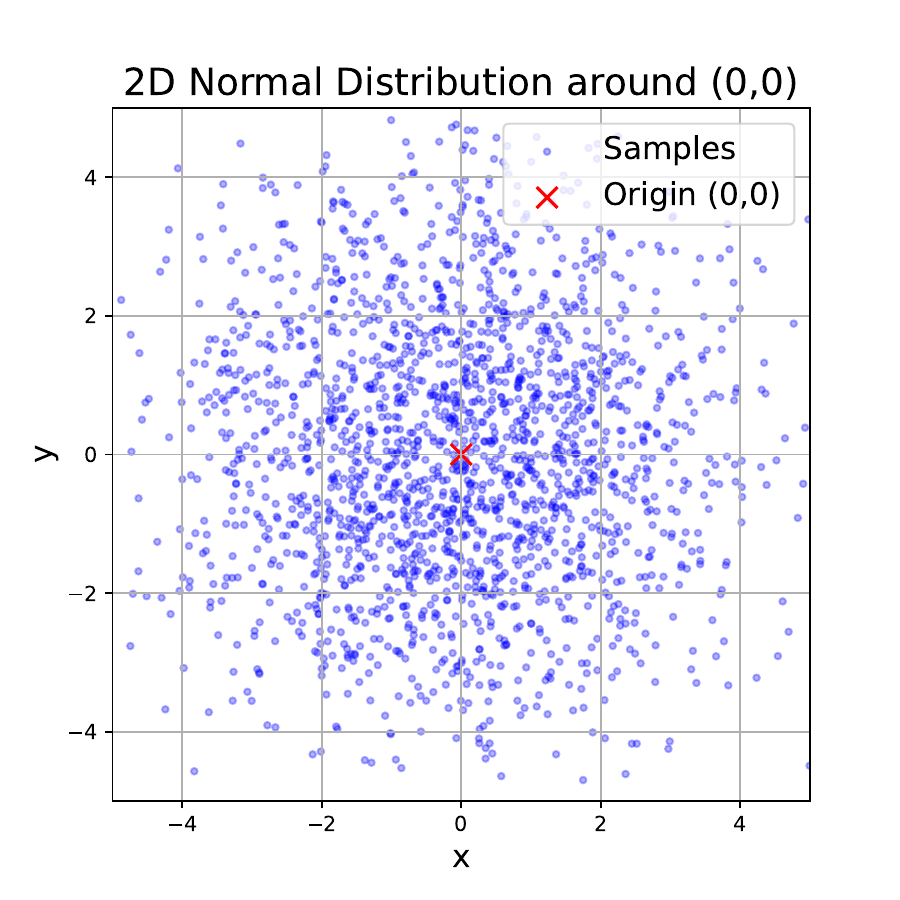}}
    \subfigure[Small scaling factor $h(\mathbf{x})$]{\includegraphics[width=0.32\linewidth]{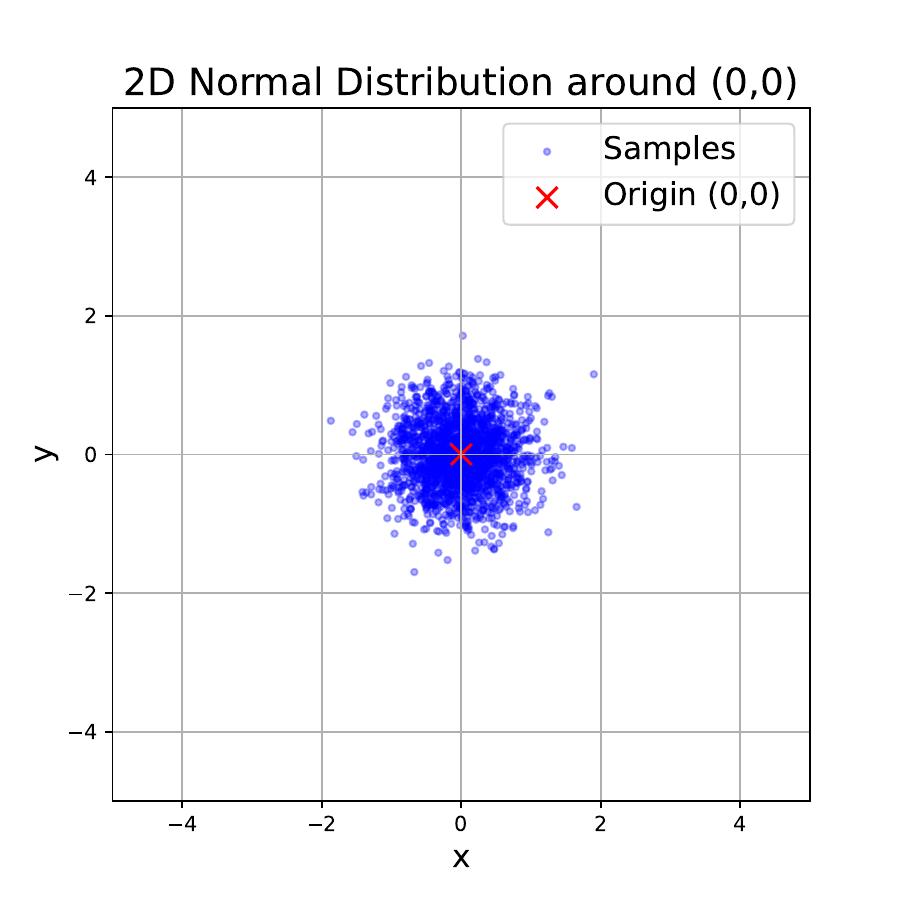}}
    \caption{Effect of a residual-dependent scaling factor $h(\mathbf{x})$ on the distribution. (a) Standard normal distribution. (b) Distribution with a large scaling factor $h(\mathbf{x})$, corresponding to a small residual $r(\mathbf{x})$, resulting in a wider spread.  (c) Distribution with a small scaling factor $h(\mathbf{x})$, corresponding to a large residual $r(\mathbf{x})$, leading to tighter local concentration.}
    \label{f.samplingdistribution}
\end{figure}

For each training point, $M$ candidate samples are generated by adding scaled Gaussian perturbations, 
\begin{equation}
\label{e.newpoint}
    \mathbf{\hat{x}}_{i_m} = \mathbf{x}_i + h(\mathbf{x}_i)\boldsymbol{\xi}_m,~\boldsymbol{\xi}_m \sim \mathcal{N}(0,I), ~m = 1,2,\dots,M.
\end{equation}
This procedure forms the candidate pool $X_{cand}=\{\mathbf{\hat{x}}_j\}_{j=1}^{NM}$.
In this way, points with larger residuals produce narrower distributions, leading to denser sampling around difficult regions,  while points with smaller residuals yield wider distributions, preventing underrepresentation of well-learned areas. Fig.~\ref{f.samplingdistribution} illustrates this effect with a 2-dimensional example.

\begin{remark}
Since the solution domain is bounded, some sampled points may fall outside the domain. To address this, we apply a reflective boundary condition, mirroring the points back into the domain.
\end{remark}

\subsection{Global Distribution-Based Selection}

To select the final collocation set from $X_{cand}$, we incorporate residual information into a residual-based weighting function $w(\mathbf{x})$. Prior work \cite{wu2023comprehensive} has constructed a global residual-based measure by explicitly evaluating residuals (Eq.~\eqref{e.residual}) over the entire domain:
\begin{equation}
\label{e.rad}
w(\mathbf{x}) = \frac{r^k(\mathbf{x})}{\mathbb{E}[r^k(\mathbf{x})]} + c,
\end{equation}
where $k \ge 0$ and $c \ge 0$ are two hyperparameters.

Numerically, the expectation $\mathbb{E}[r(\mathbf{x})]$ is approximated via Monte Carlo integration using a dense set of points $X_{dense} = \{\mathbf{\tilde{x}}_t\}_{t=1}^S$ ($S \gg N$). Each point is then assigned an unnormalized weight
\begin{equation}
\label{e.weightfun}
w(\mathbf{\tilde{x}}_t) = \frac{r^k(\mathbf{\tilde{x}}_t)}{\frac{1}{S}\sum_{j=1}^S r^k(\mathbf{\tilde{x}}_j)} + c,
\end{equation}
which is subsequently normalized to form a discrete probability mass function (PMF) for sampling $N$ collocation points. However, this procedure is computationally expensive because residual evaluation often involves second- or higher-order derivatives, which require significant resources, especially in high-dimensional settings.  

To reduce cost, we propose a lightweight approximation in which each candidate point $\mathbf{\hat{x}}_{i_m}$ sampled around $\mathbf{x}_i$ inherits the residual value of its anchor point $\mathbf{x}_i$, i.e.,
\begin{equation}
\label{e.assia}
r(\mathbf{\hat{x}}_{i_m}) = r(\mathbf{x}_i), \quad m = 1,2,\dots,M.
\end{equation}
This enables us to obtain weights without recomputing residuals for every candidate. The weights are then normalized as
\begin{equation}
\label{e.pmf}
p(\mathbf{\hat{x}}_j) = \frac{w(\mathbf{\hat{x}}_j)}{\sum_{l=1}^{NM} w(\mathbf{\hat{x}}_l)},
\end{equation}
and $N$ new samples are drawn from this distribution to form the updated training set.

\begin{figure}[!h]
    \centering
    \subfigure[Sparse training points]{\includegraphics[width=0.45\linewidth]{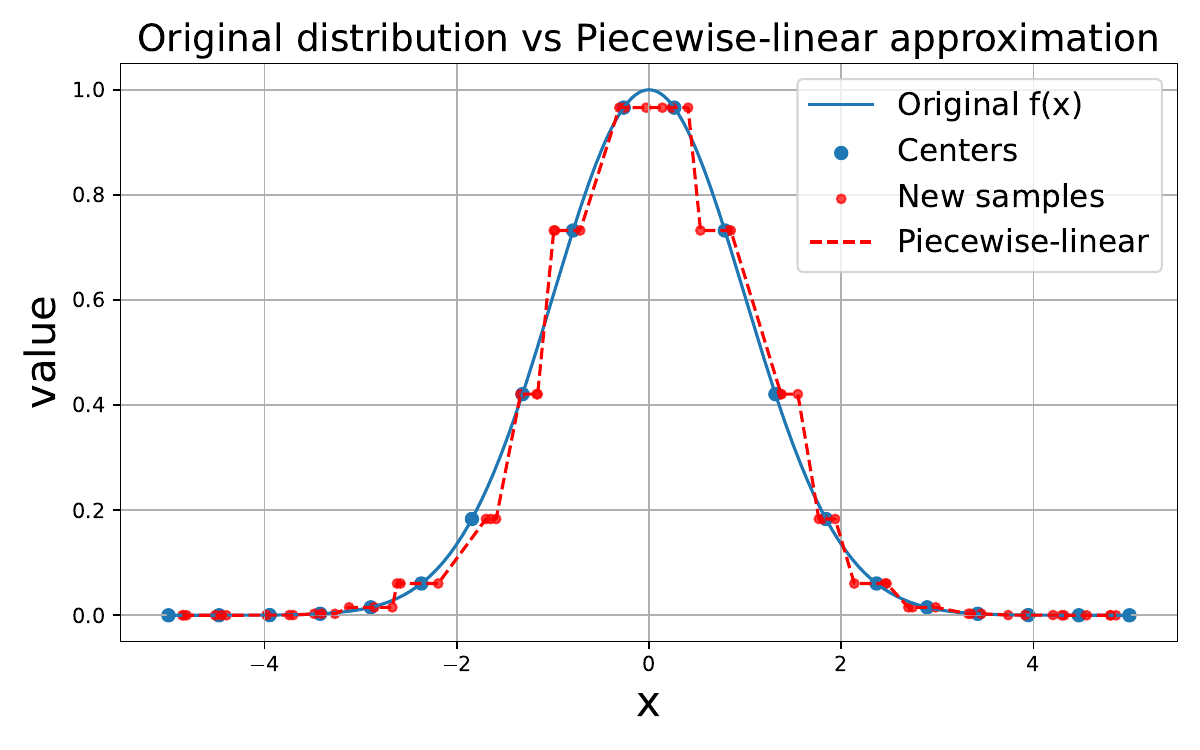}}
    \subfigure[Dense training points]{\includegraphics[width=0.45\linewidth]{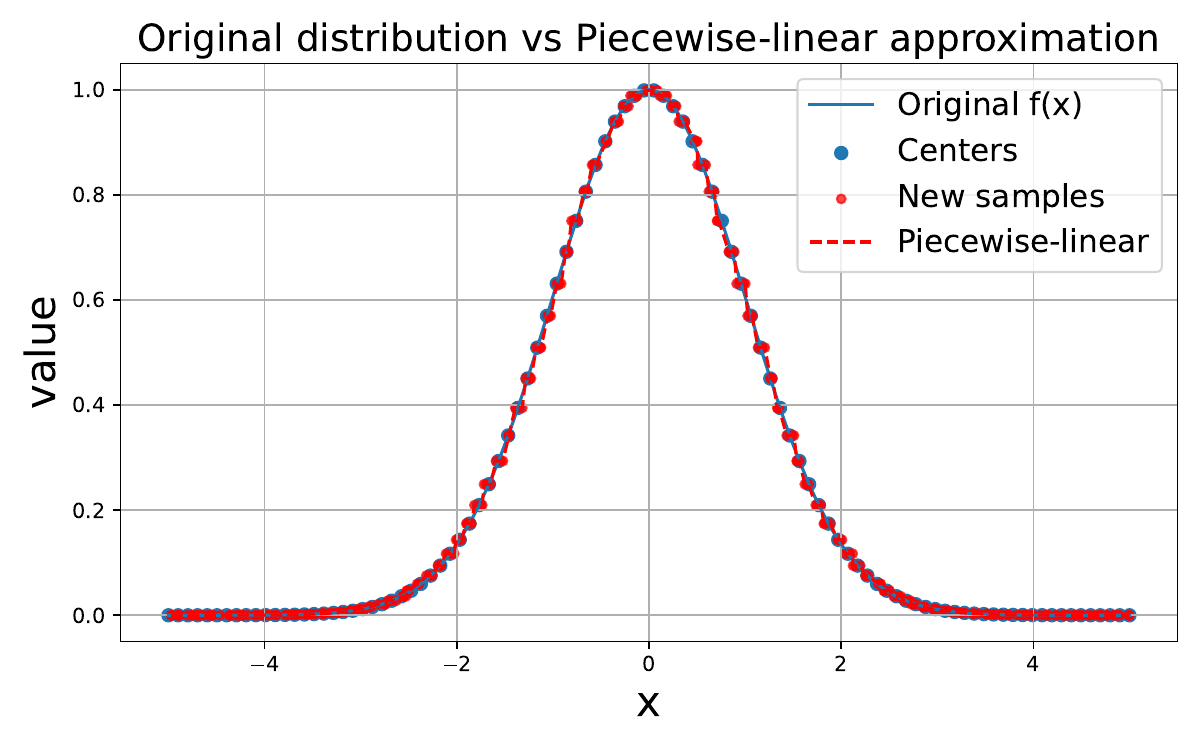}}
    \caption{Piecewise-linear approximation with different numbers of training points.  
    (a) With sparse training points, the approximation is coarse and deviates from the original distribution.  
    (b) With dense training points, the approximation becomes finer and closely aligns with the original curve.}
    \label{f.approximatedistribution}
\end{figure}

Fig.~\ref{f.approximatedistribution} provides a visualization of the piecewise-linear approximation of the 1-dimensional residual-based distribution. In the first case, only a few training points are available, and the approximation takes the form of a coarse piecewise-linear curve, which deviates noticeably from the true distribution. When more training points are used, the piecewise-linear approximation becomes much finer and closely matches the original curve. 
This illustrates that the piecewise-linear approximation remains effective: with few training points it provides a reasonable estimate of the distribution, and with more points it rapidly converges to the true distribution, confirming its validity as a lightweight alternative to costly residual-based computations.

\subsection{Algorithm}
The complete procedure is summarized in Algorithm~\ref{alg.glf}. The stopping criterion can be either a fixed number of iterations or the satisfaction of a convergence threshold.  

\begin{algorithm}
\caption{Global-Local Fusion Sampling (GLF)}
\label{alg.glf}
\begin{algorithmic}[1]
\State Initialize with $N$ training points
\State Train the PINN for a fixed number of iterations
\While{stopping criterion not met}
    \State Evaluate residuals $r(\mathbf{x}_i)$ using Eq.\eqref{e.residual}
    \State Generate candidate samples using Eq.\eqref{e.newpoint}
    \State Assign residuals to candidates and construct $p(\mathbf{x})$ via Eq.\eqref{e.assia} and \eqref{e.pmf}
    \State Sample $N$ points from $p(\mathbf{x})$ to update the training set
    \State Train the PINN for a fixed number of iterations
\EndWhile
\end{algorithmic}
\end{algorithm}

\section{Experiments}
\label{s:exps}

\subsection{Experimental Setup}
\paragraph{Baselines}
We make comparisons with a global method RAD \cite{wu2023comprehensive} and a local method RSmote \cite{luo2025imbalanced}.

\paragraph{Benchmarks} 
We evaluate our proposed Global--Local Fusion (GLF) sampling strategy on five representative PDE benchmarks, the first three are 2-dimensional and the rest are high-dimensional.
\begin{itemize}
    \item Allen-Cahn Equation:
    \begin{equation}
    \label{e.allen}
    \left\{\begin{aligned}
    &u_t = 0.001 u_{xx} + 5(u-u^3), \\
    &u(x, 0) = x^2\cos(\pi x), \\
    &u(-1, t) = u(1, t) = -1,  \\
   & x\in[-1, 1], t\in[0,1]. 
    \end{aligned} 
    \right.
    \end{equation}
    
    \item Burgers’ Equation:
    \begin{equation}
    \label{e.burges}
    \left\{\begin{aligned}
   & u_t + u u_x = \frac{1}{100\pi}u_{xx}, \\
    &u(x, 0) = -\sin(\pi x), \\
    &u(-1, t) = u(1, t) = 0,  \\
    &x\in[-1, 1], t\in[0,1]. 
    \end{aligned} 
    \right. 
    \end{equation}
    
    \item Laplace Equation:
    \begin{equation}
    \label{e.laplace}
     \left\{\begin{aligned}
   & r\frac{dy}{dr} + r^2\frac{dy^2}{dr^2} + \frac{dy^2}{d\theta^2} = 0, \\
    &y(1,\theta) = \cos(\theta), \\
    &y(r, \theta +2\pi) = y(r, \theta),  \\
    &r \in [0, 1],  \theta \in [0, 2\pi].
       \end{aligned} 
    \right. 
    \end{equation}
    The analytical  solution is $y=r\cos(\theta).$
    
    \item 10-dimensional Dispersive Equation:
    \begin{equation}
    \label{e.highdimkdv}
    \left\{\begin{array}{l}
    u_t + \sum \limits_{i=1}^{d}\partial_{x_i}^{3}u = f(\mathbf{x}, t),~~ \mathbf{x}\in \Omega \\
    u(\mathbf{x}, 0) = \sin(\frac{1}{d}\sum \limits_{i=1}^{d}x_i), \\
    x\in \Omega=[-1, 1]^d, t\in[0,1],
    \end{array}\right. 
    \end{equation}
    where $f(\mathbf{x}, t)=-\frac{1}{d}[\sin(\frac{1}{d}\sum \limits_{i=1}^{d}x_i)+\cos(\frac{1}{d}\sum \limits_{i=1}^{d}x_i)]\exp(\frac{-t}{d^2})$, and $d=10$ is the dimension, which admits the exact solution $u(\mathbf{x})=\sin(\frac{1}{d}\sum \limits_{i=1}^{d}x_i)\exp(\frac{-t}{d^2})$.
    
    \item 20-dimensional Reaction-Diffusion Equation:
    \begin{equation}
    \label{e.highdimreaction}
    \left\{\begin{array}{l}
    u_t = \bigtriangleup u - 0.2u - de^{-0.2t},~~ \mathbf{x}\in \Omega \\
    u(\mathbf{x}, 0) = \|\mathbf{x}\|^2/2, \\
    x\in \Omega=[-1, 1]^d, t\in [0,1]
    \end{array}\right. 
    \end{equation}
    where $d=20$ is the dimension. The analytical solution is $u(\mathbf{x}, t) = \frac{1}{2}\|\mathbf{x}\|^2 e^{-0.2t}$.
\end{itemize}

\paragraph{Implementation details}
For GLF, the number $M$ of candidate points per neighborhood is set to 3, $\epsilon$ is fixed at $1 \times 10^{-8}$, and $\alpha$ is set to $1 \times 10^{-4}$ for the first three low-dimensional examples and $1 \times 10^{-3}$ for the last two high-dimensional examples.
For RAD, the extra number of points is 100000, and for RSmote, the ratio is set to 0.45. 
In each sampling iteration, we first use the Adam optimizer for 1000 steps and then switch to the L-BFGS optimizer for another 1000 steps to train the model. Each experiment is run 5 times with diﬀerent random seeds, and the mean and standard deviation of the errors are computed.
All methods are trained for 100 sampling iterations, which also serves as the termination condition.
In all examples, the hyperbolic tangent (tanh) is selected as the activation function. 
Table \ref{T.param} summarizes the network architecture used for each example.
All the experiments are conducted on a workstation equipped with an NVIDIA-3090 GPU.
\begin{table}[h]
\centering
\caption{Summary of the experimental setup.}
\begin{adjustbox}{width=\textwidth}
\begin{tabular}{lcccc}
\toprule[2pt]
 Problems & Depth & Width & Dimension (d) & \#Training points \\
\midrule[1pt]
 Allen-Cahn & 3 & 64 & 2 & 2000 \\
 Burgers & 3 & 64 & 2 & 2000\\
 Laplace & 3 & 20 & 2 & 2000\\
 Dispersive & 3 & 2d & 10 & 10000\\
 Reaction-Diffusion & 3 & 2d & 20 & 10000\\
 \bottomrule[2pt]
\end{tabular}
\end{adjustbox}
\label{T.param}
\end{table}

\paragraph{Evaluation metrices}
We use the $L_2$ relative error to evaluate the performance, which is defined as $\frac{\|\hat{u}-u\|_2}{\|u\|_2}$, and the GPU memory (MB) requirements for each method.

\subsection{Comparison with Global and Local Methods}
\label{s:comparison}

\begin{figure}[!ht]
\centering
    \subfigure[Allen-Cahn]{\includegraphics[width=0.45\linewidth]{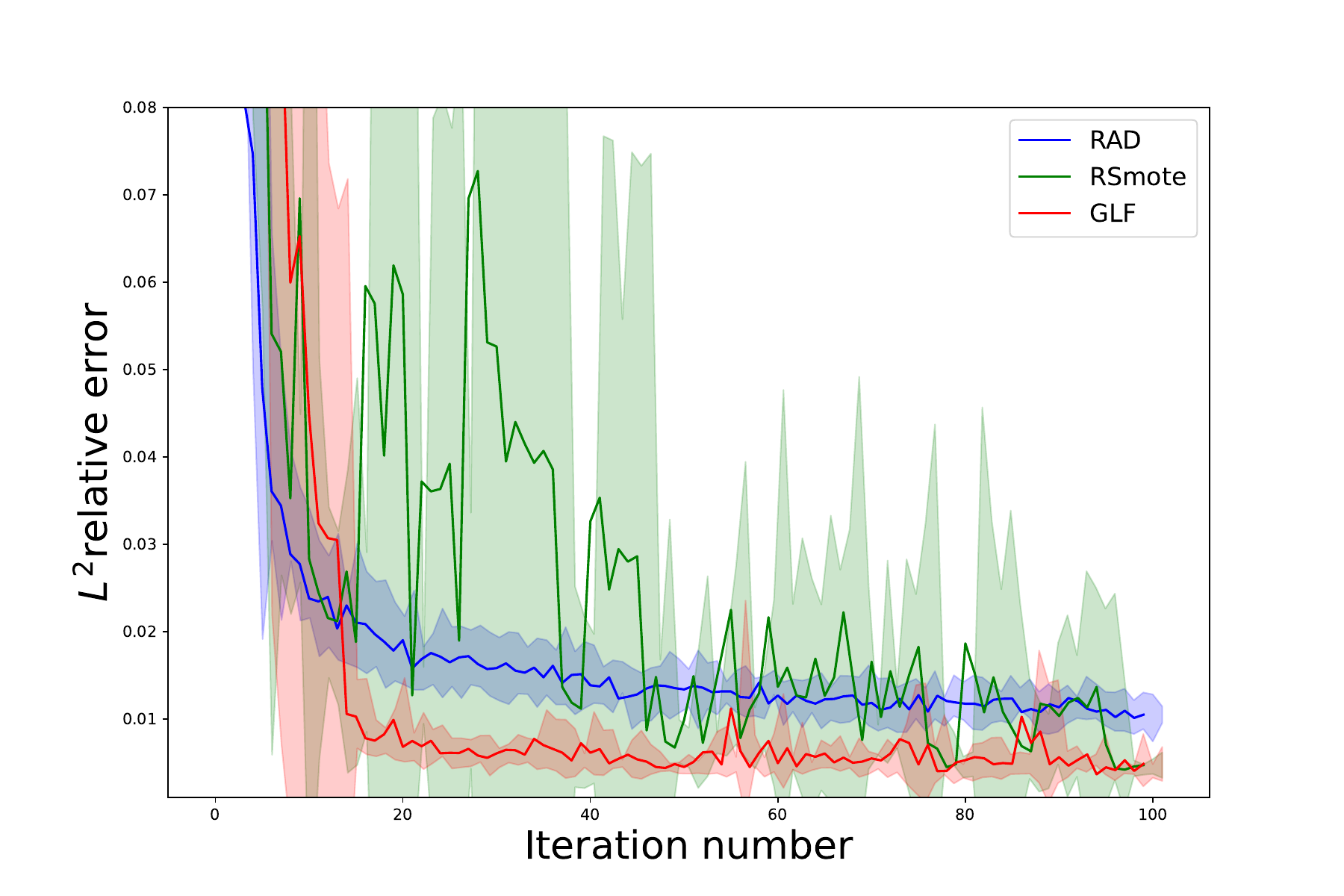}}
    \subfigure[Burgers]{\includegraphics[width=0.45\linewidth]{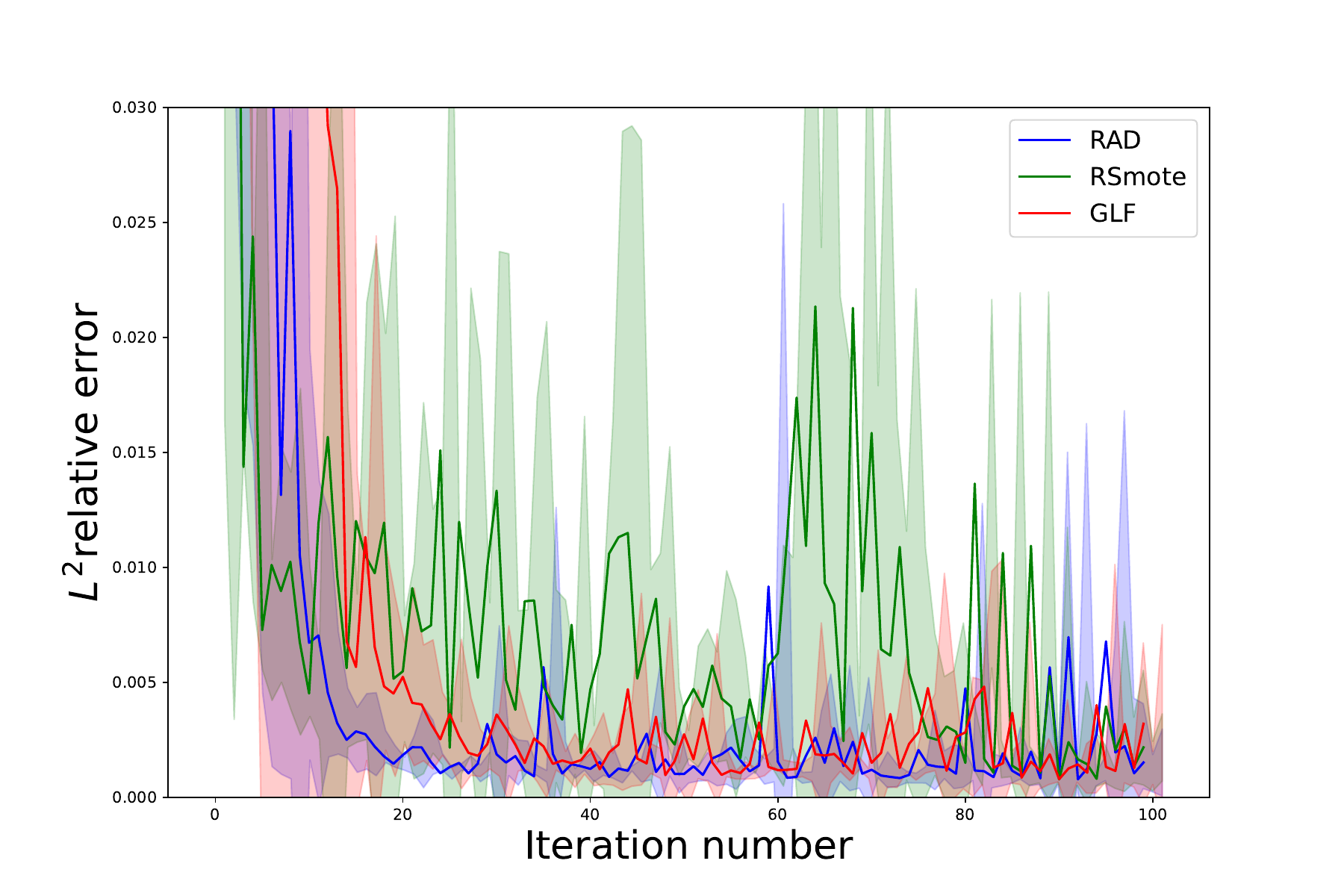}}
    \subfigure[Laplace]{\includegraphics[width=0.45\linewidth]{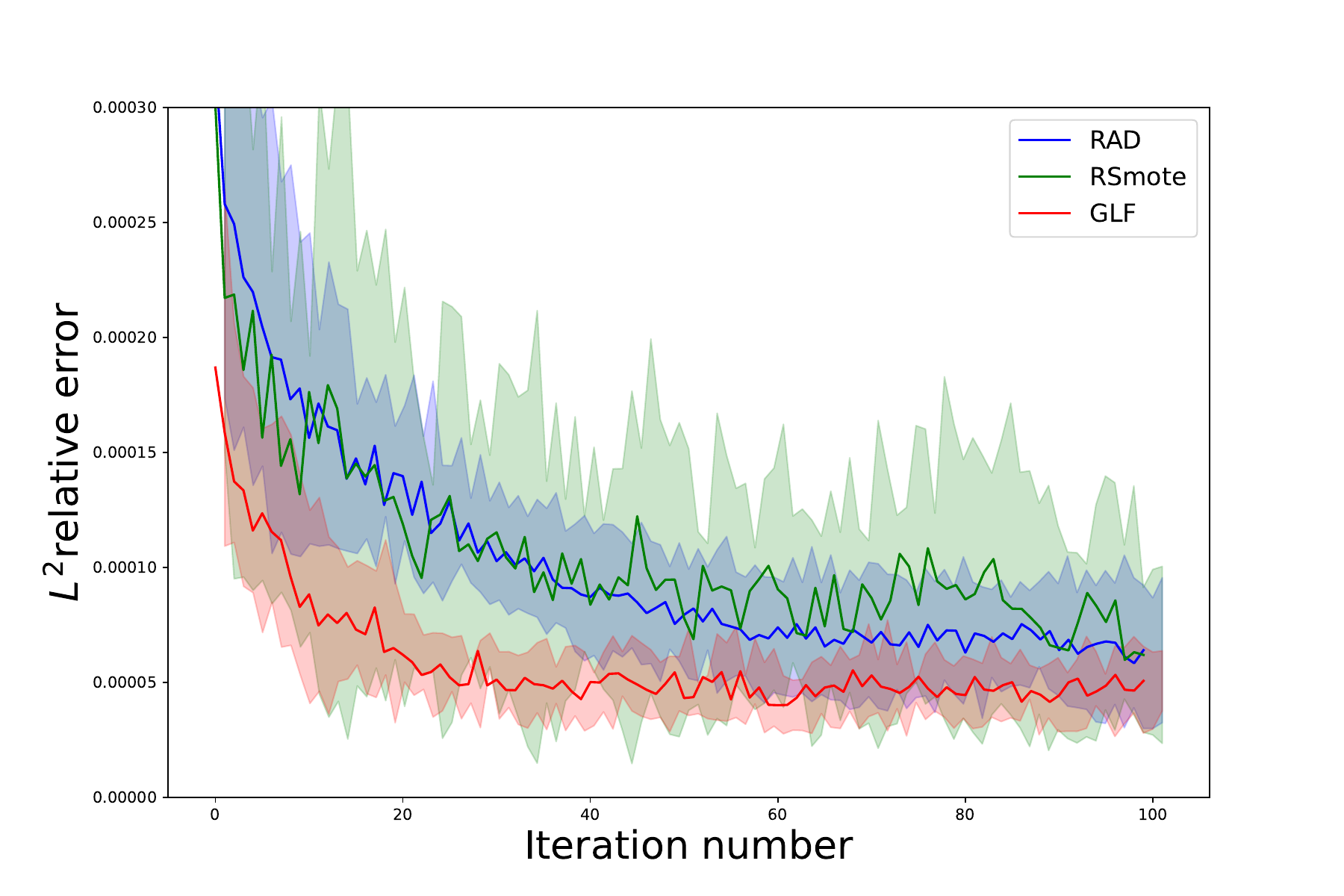}}
    \subfigure[Dispersive]{\includegraphics[width=0.45\linewidth]{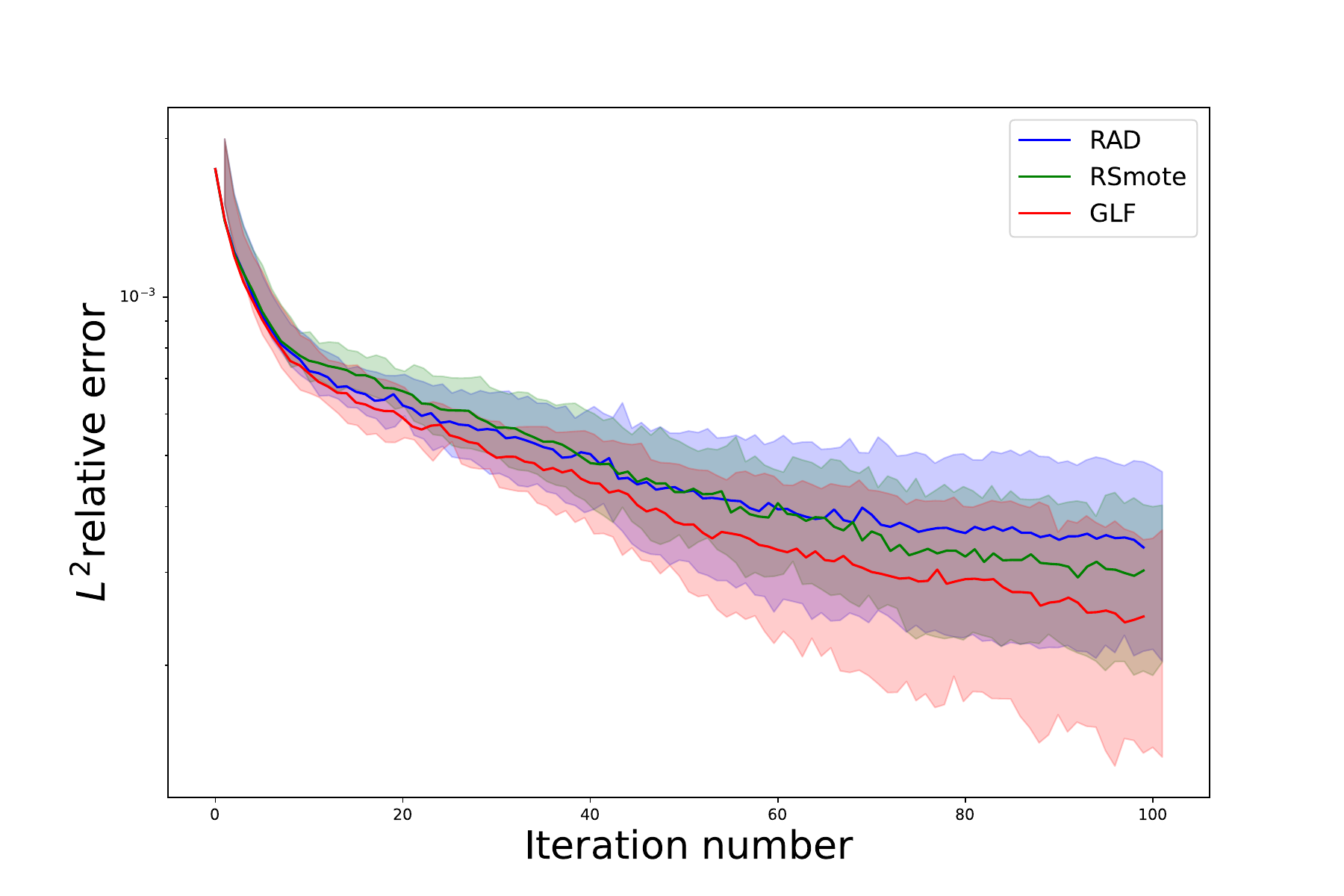}}
    \subfigure[Reaction-Diffusion]{\includegraphics[width=0.45\linewidth]{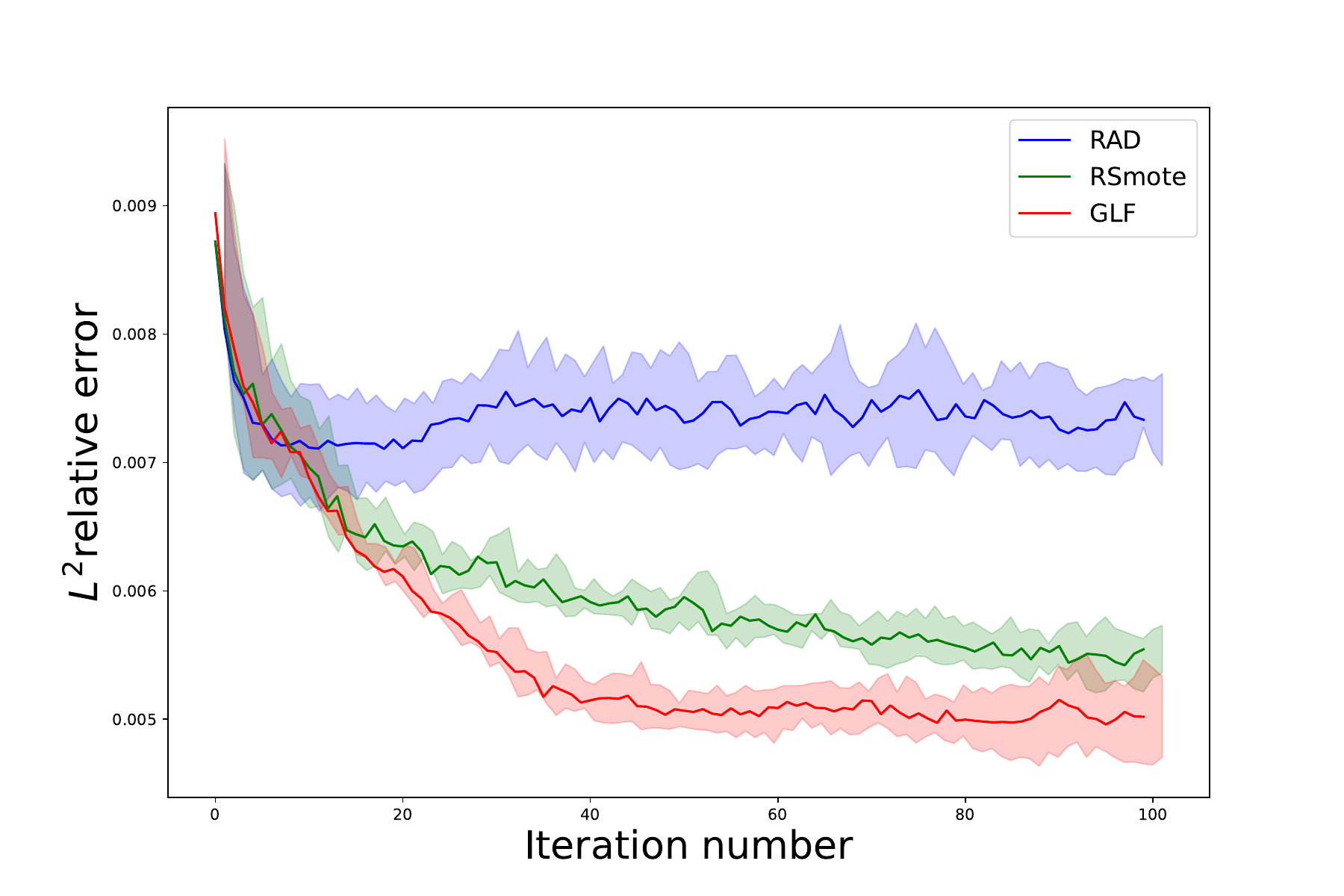}}
\caption{Training loss curves for different PDEs. Red line: Mean values of GLF method; Blue line: Mean values of RAD-100000 method; Green line: Mean values of RSmote method. The shaded areas represent the corresponding standard deviations.}
\label{f.loss_curves}
\end{figure}

Here we compare the proposed GLF with the global method RAD and the local method RSmote across five PDE benchmarks. Table~\ref{tab:pde_summary} reports the quantitative results while Fig.~\ref{f.loss_curves} shows representative training loss curves. Together, they provide a comprehensive view of both the final performance and the training dynamics.

On accuracy, GLF consistently achieves the lowest or near-lowest relative error. For Allen–Cahn, Laplace, Dispersive, and Reaction–Diffusion equations, GLF outperforms both RAD and RSmote by a clear margin, demonstrating its ability to capture complex solution structures more effectively. On Burgers’ equation, GLF achieves accuracy comparable to RSmote and superior to RAD. These results indicate that GLF more reliably captures complex solution structure across heterogeneous PDE families, yielding consistently higher fidelity reconstructions.

On efficiency, GLF matches the lightweight memory usage of RSmote and substantially improves over RAD. For the dispersive and reaction-diffusion equations, GLF uses approximately a quarter of the RAD GPU memory while achieving better accuracy. This efficiency stems from GLF’s neighborhood–residual assignment, which avoid computing extra residuals on a dense set, thereby reducing per-iteration overhead.

On stability, the loss curves in Fig.~\ref{f.loss_curves} reveal the underlying reason for GLF’s superior performance. RSmote, which relies solely on local residual refinement, often misjudges which regions require additional sampling. This leads to oscillations in the loss during the data updating process, reflecting unstable training behavior, especially the complicated systems (see Fig.~\ref{f.loss_curves} (a) and (b)). By contrast, RAD maintains stable convergence but at a much higher computational cost. 
GLF inherits the global stability of RAD while avoiding its overhead. The trajectories are smoother, settle faster, and converge to lower loss values, reflecting a more stable training process.

\begin{figure}[!h]
    \centering
    \subfigure[RAD solution]{\includegraphics[width=0.32\linewidth]{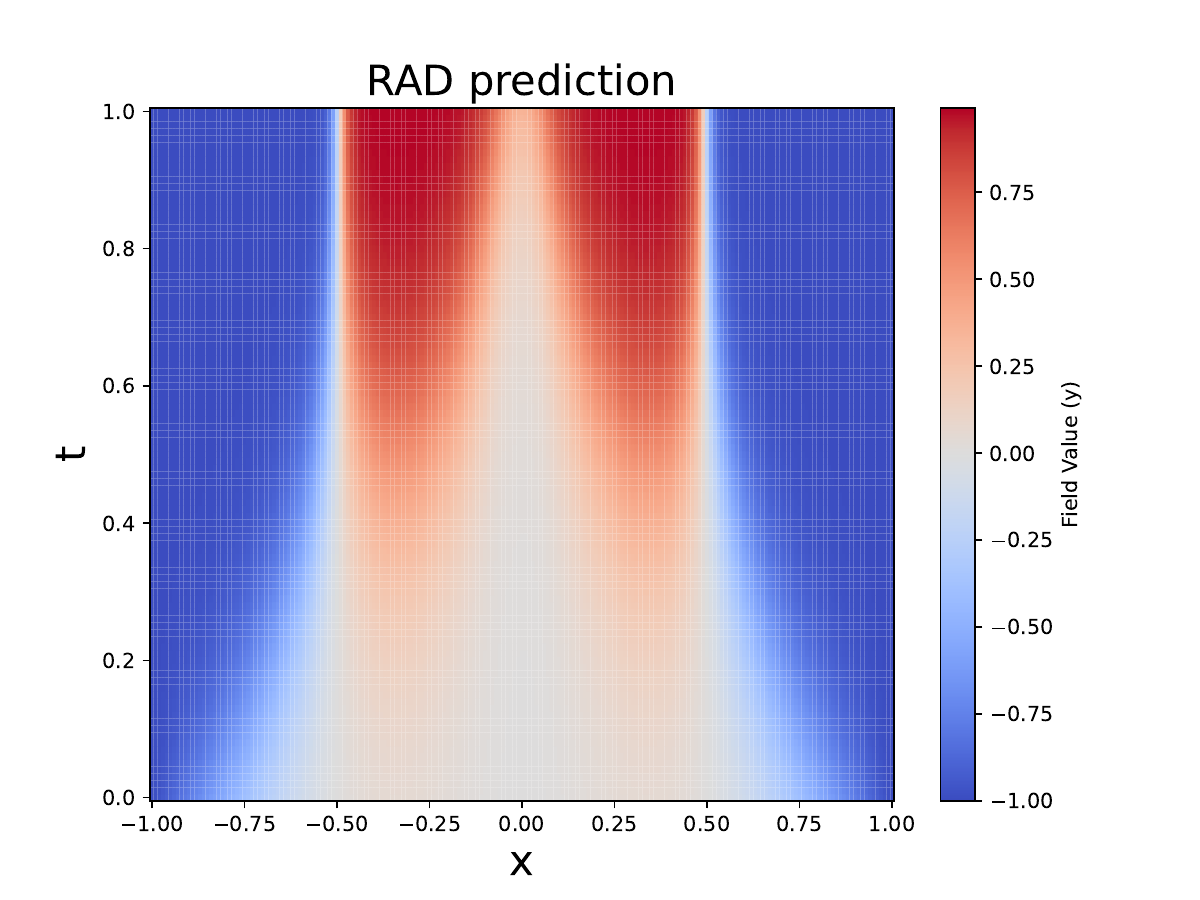}}
    \subfigure[RSmote solution]{\includegraphics[width=0.32\linewidth]{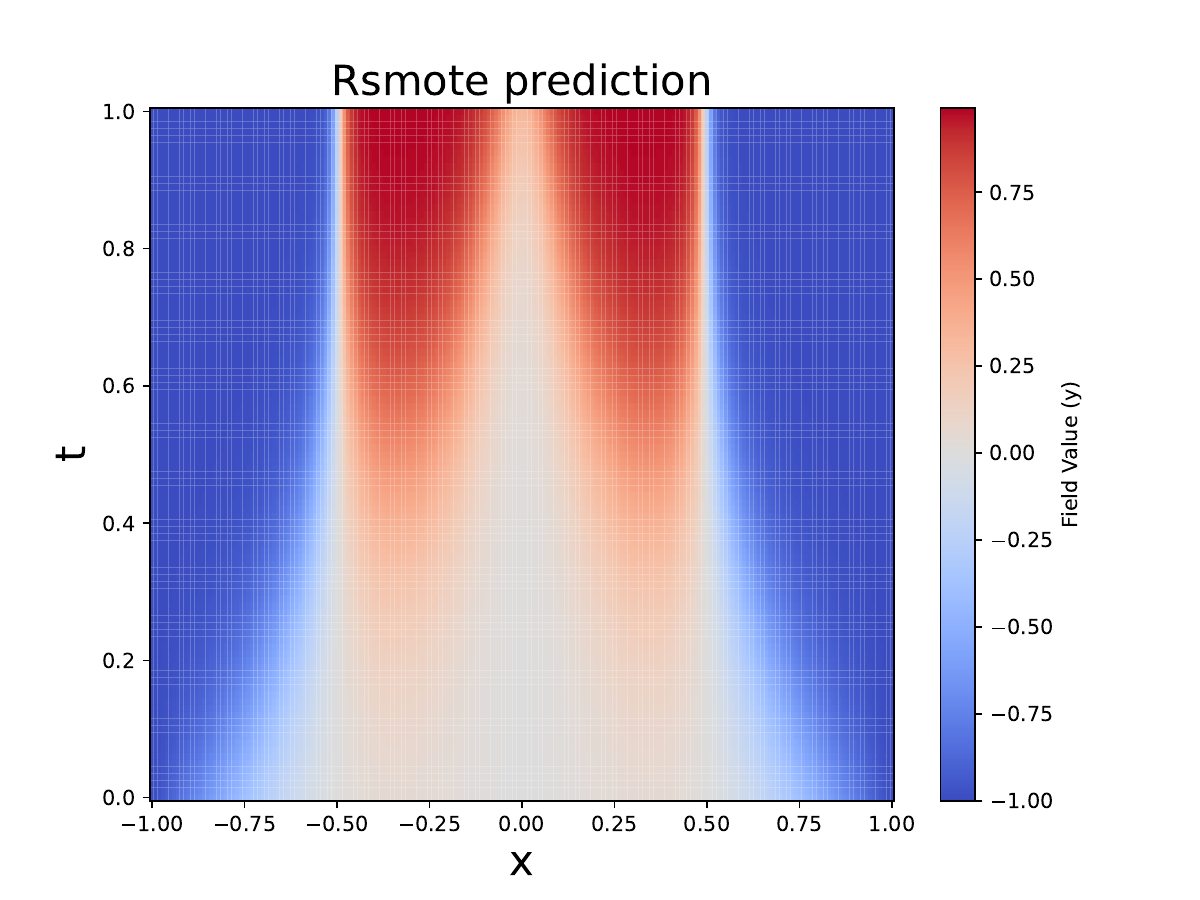}}
    \subfigure[GLF solution]{\includegraphics[width=0.32\linewidth]{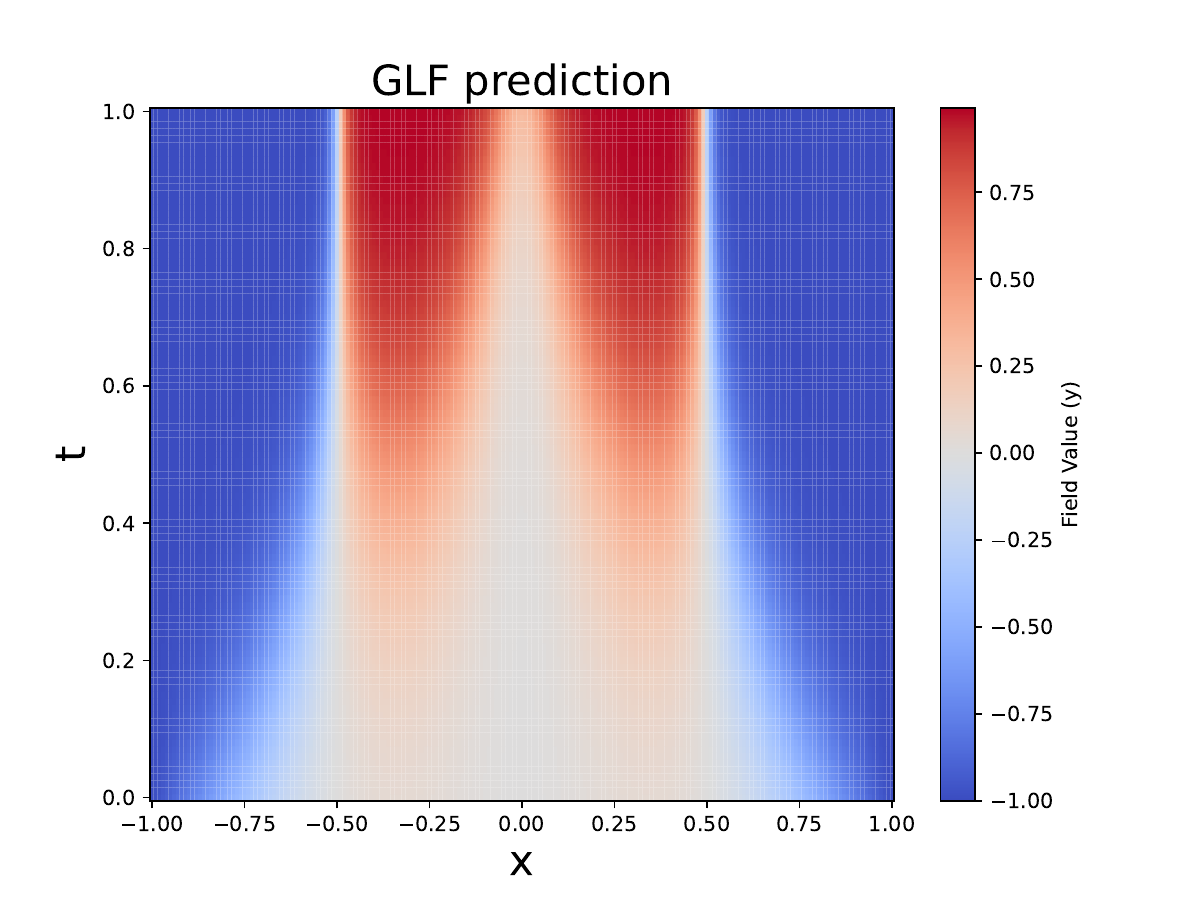}}
    \subfigure[RAD difference]{\includegraphics[width=0.32\linewidth]{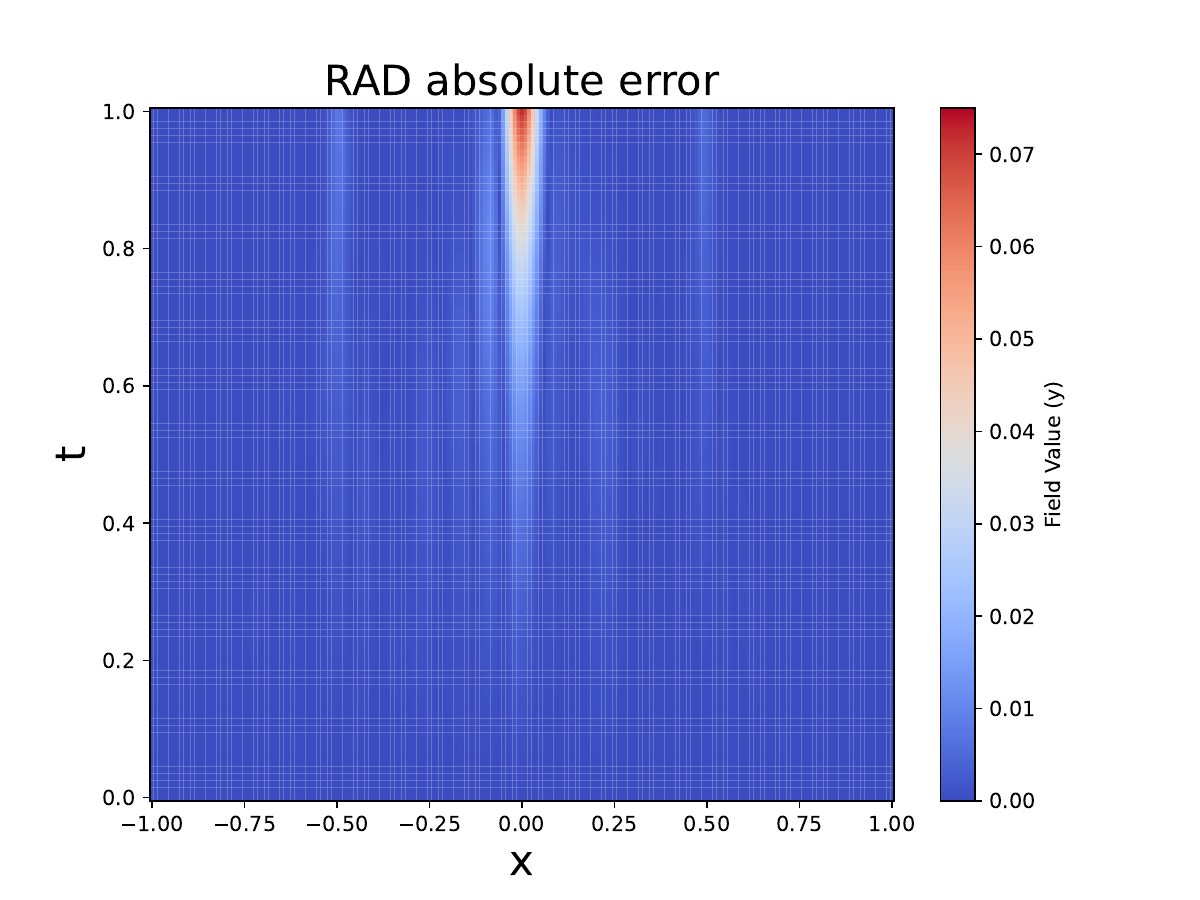}}
    \subfigure[RSmote difference]{\includegraphics[width=0.32\linewidth]{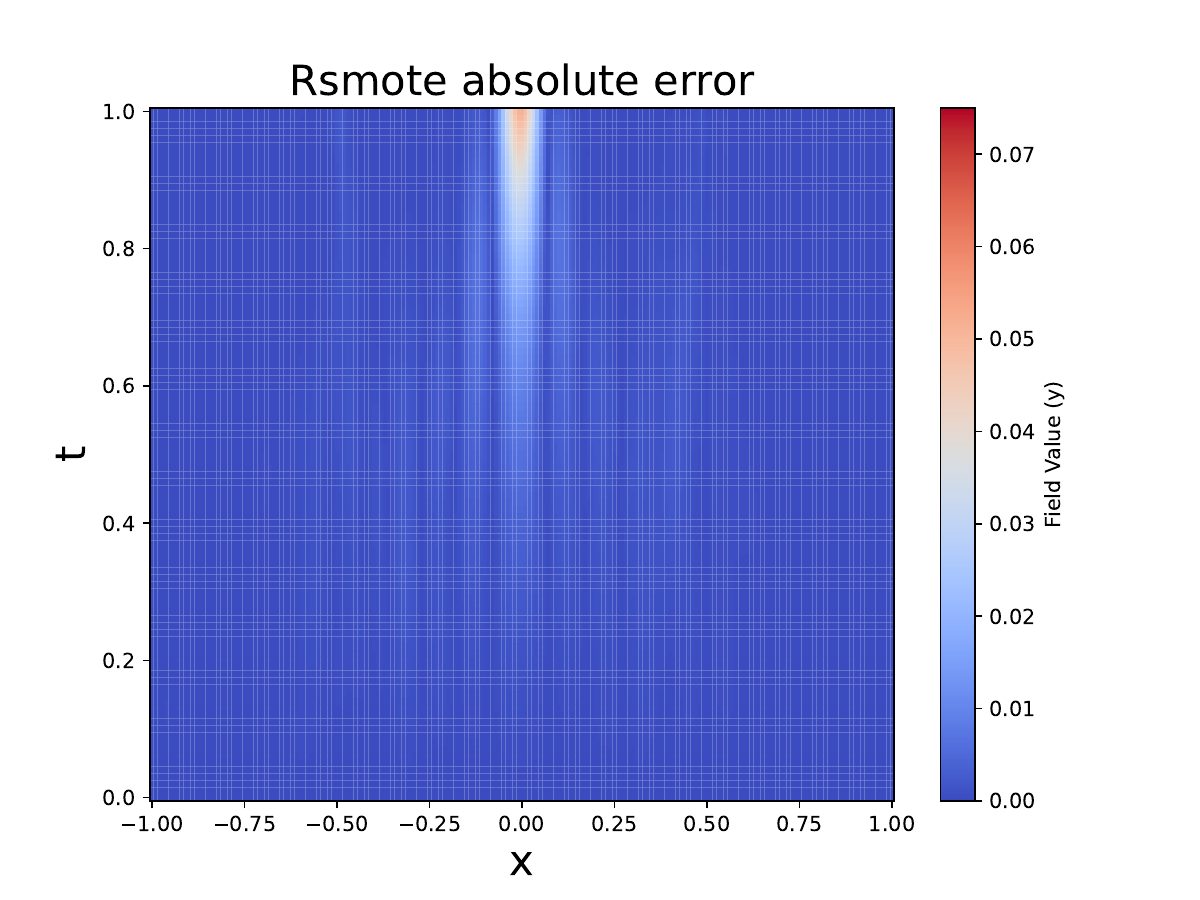}}
    \subfigure[GLF difference]{\includegraphics[width=0.32\linewidth]{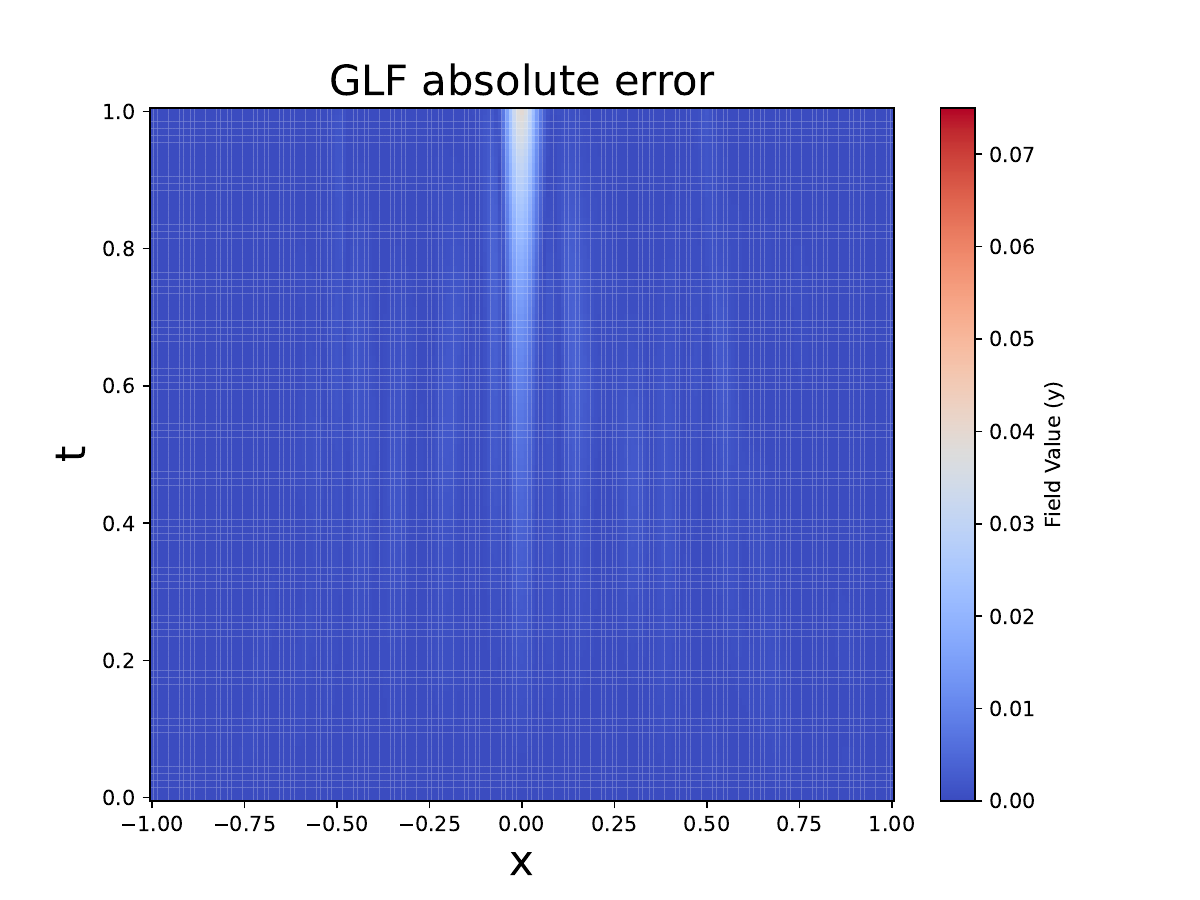}}
    \subfigure[RAD distribution]{\includegraphics[width=0.32\linewidth]{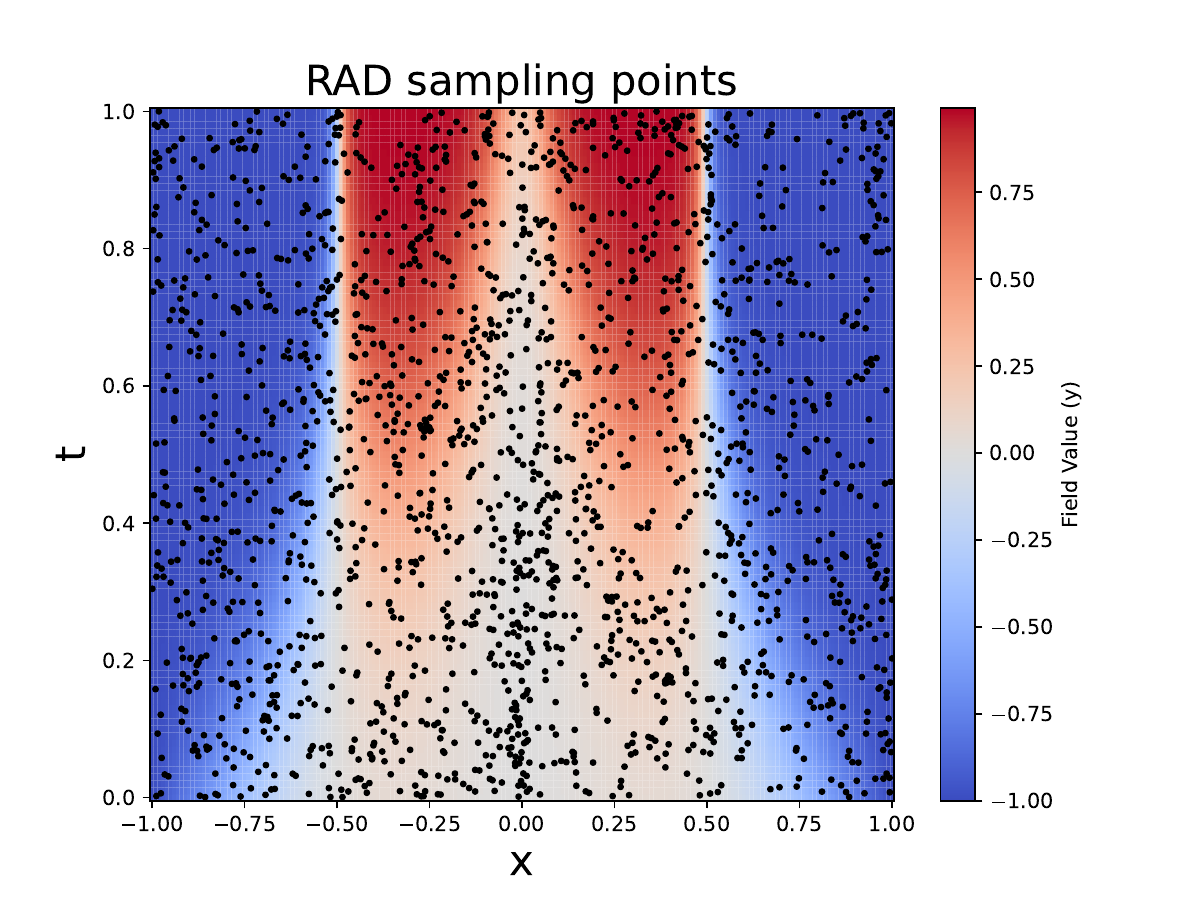}}
    \subfigure[RSmote distribution]{\includegraphics[width=0.32\linewidth]{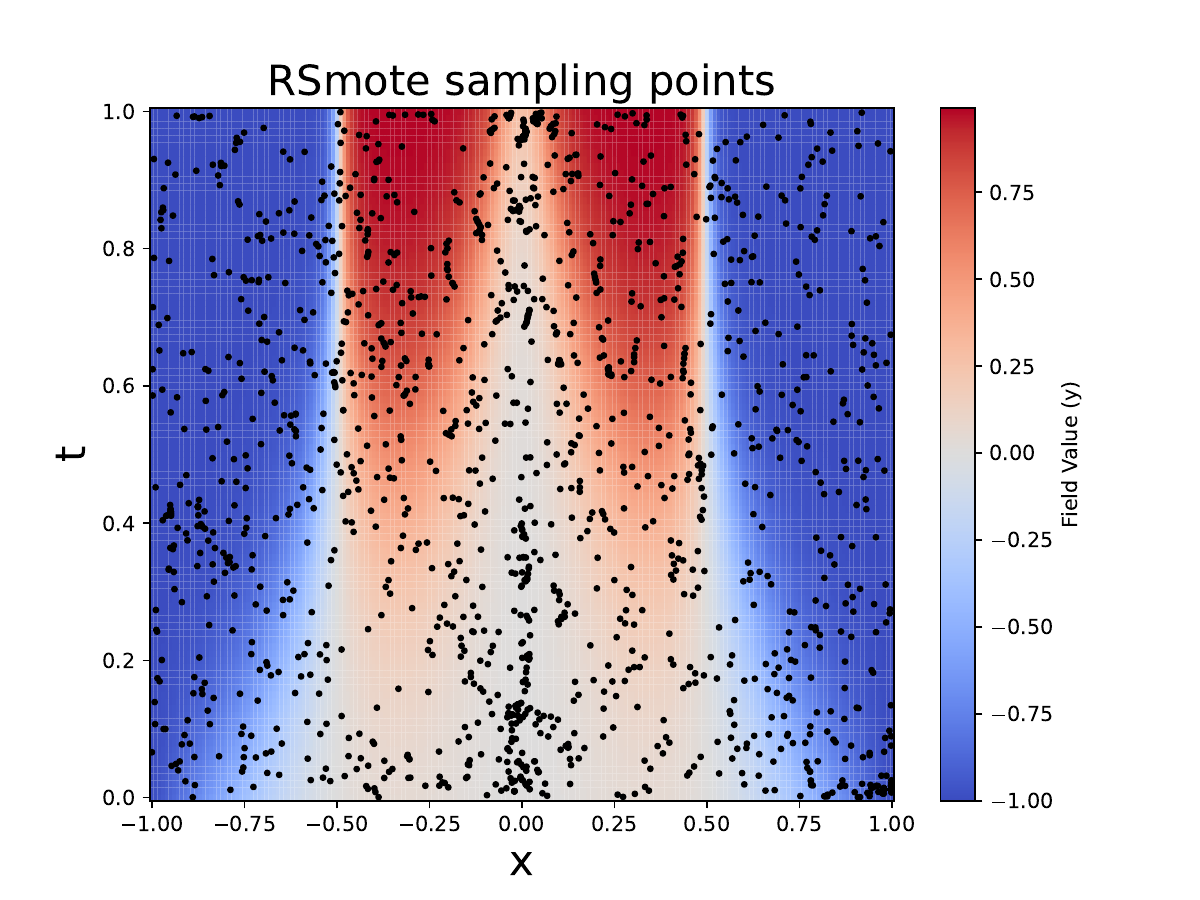}}
    \subfigure[GLF distribution]{\includegraphics[width=0.32\linewidth]{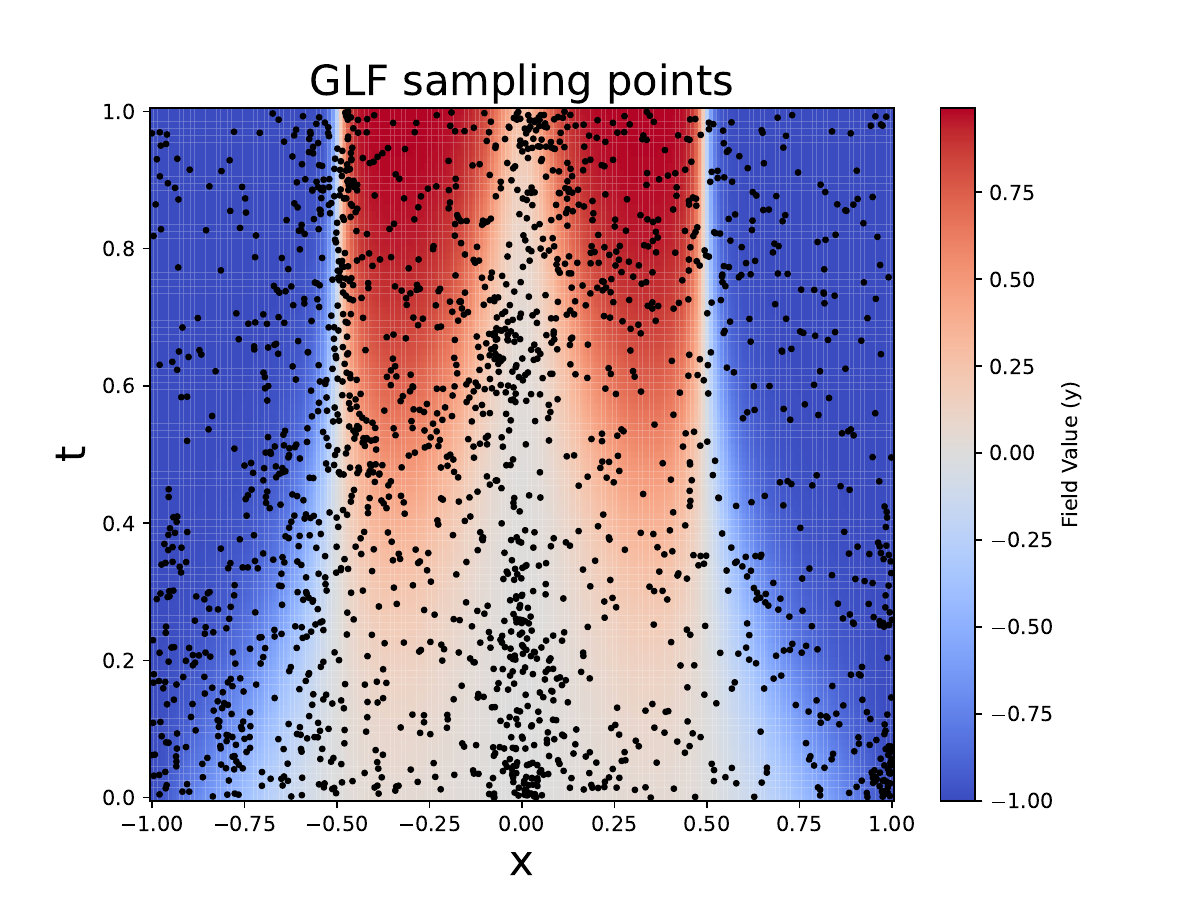}}
    \caption{Solution fields for Allen-Cahn Equation. (a)-(c): RAD solution, RSmote solution and GLF solution; (d)-(f): Absolute differences; (g)-(i): The distribution of the final points. }
    \label{f.allen_field}
\end{figure}

\begin{figure}[!h]
    \centering
    \subfigure[RAD solution]{\includegraphics[width=0.32\linewidth]{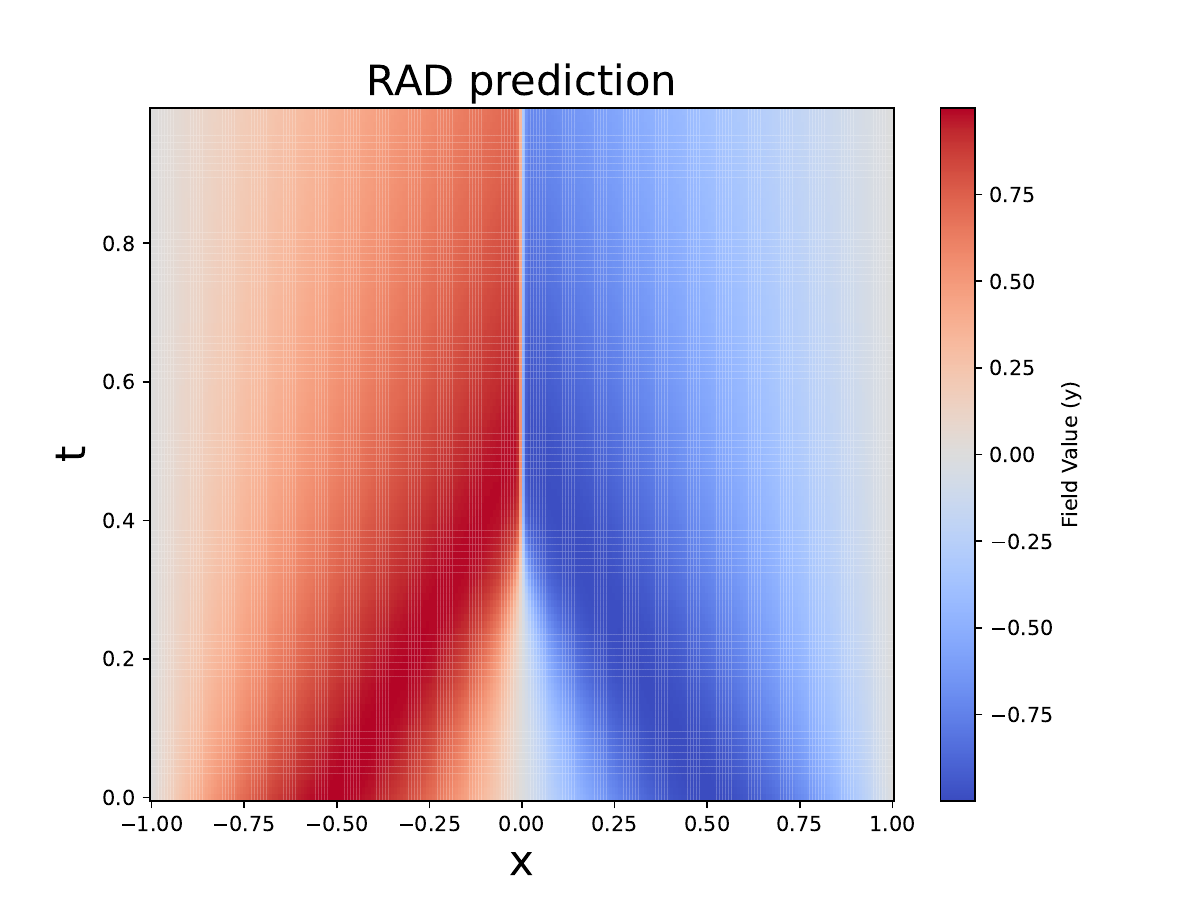}}
    \subfigure[RSmote solution]{\includegraphics[width=0.32\linewidth]{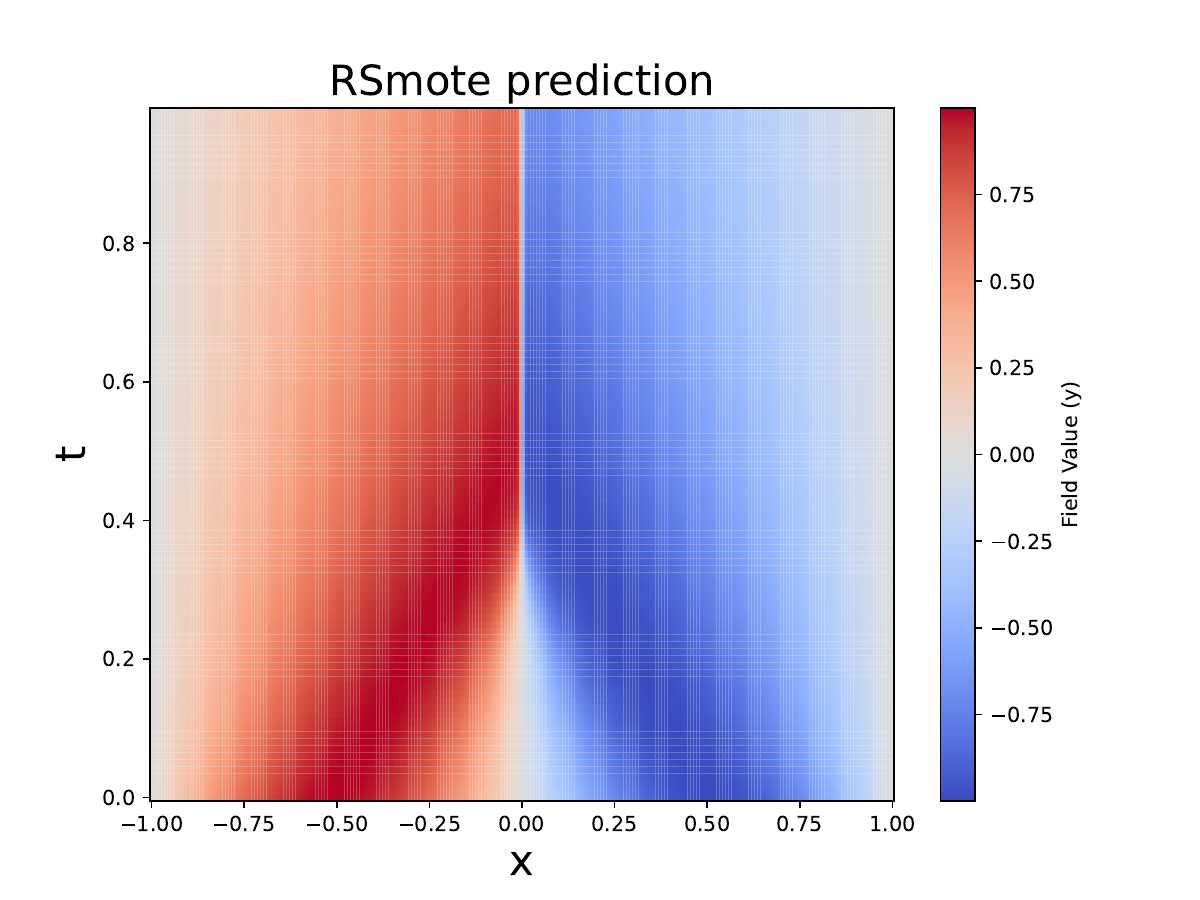}}
    \subfigure[GLF solution]{\includegraphics[width=0.32\linewidth]{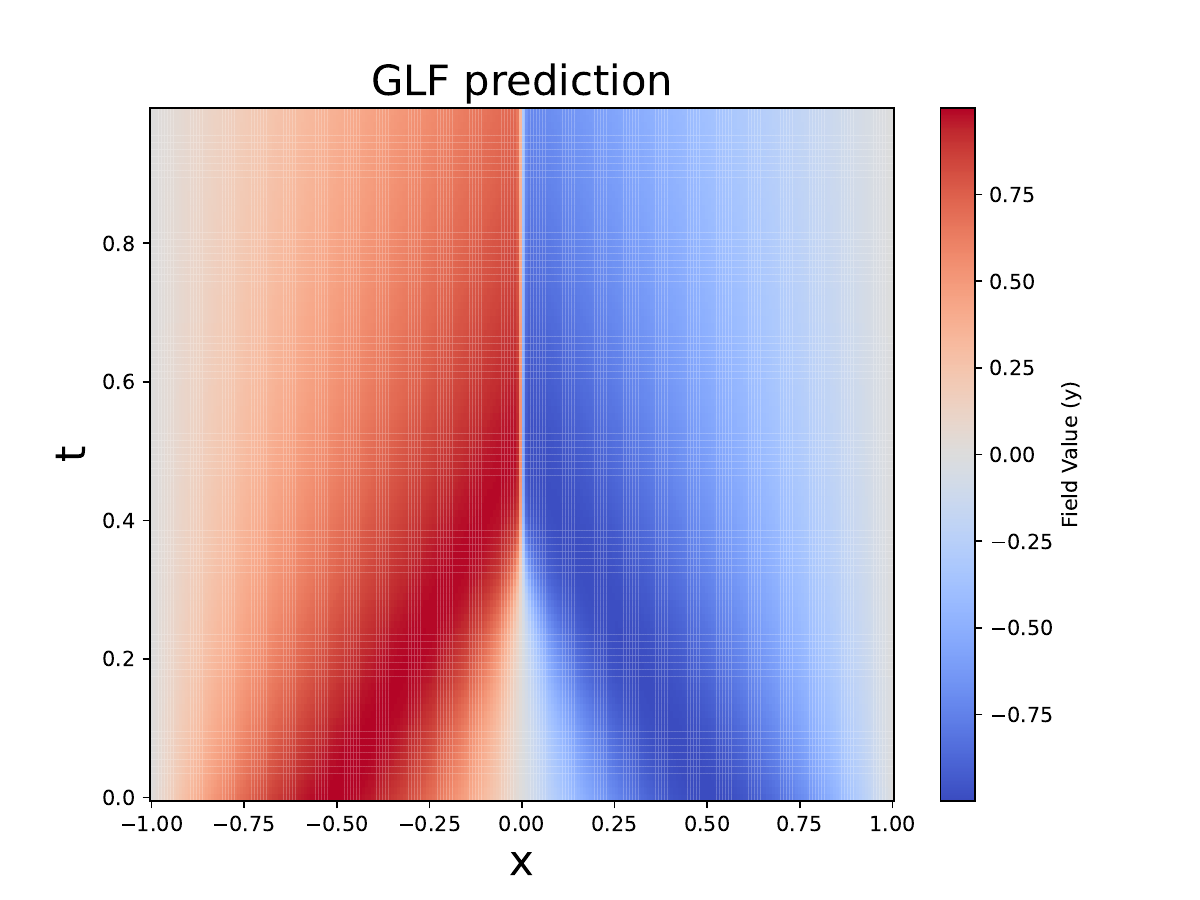}}
    \subfigure[RAD difference]{\includegraphics[width=0.32\linewidth]{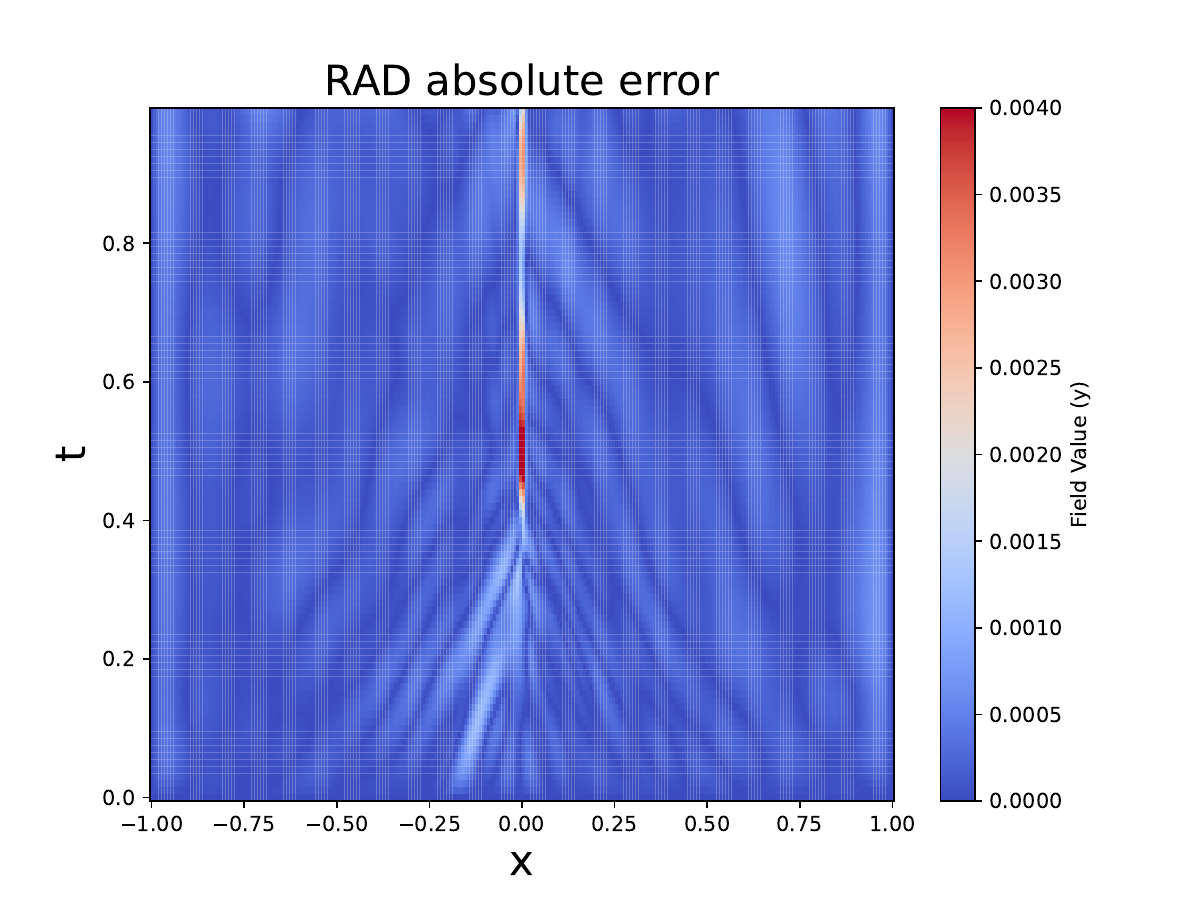}}
    \subfigure[RSmote difference]{\includegraphics[width=0.32\linewidth]{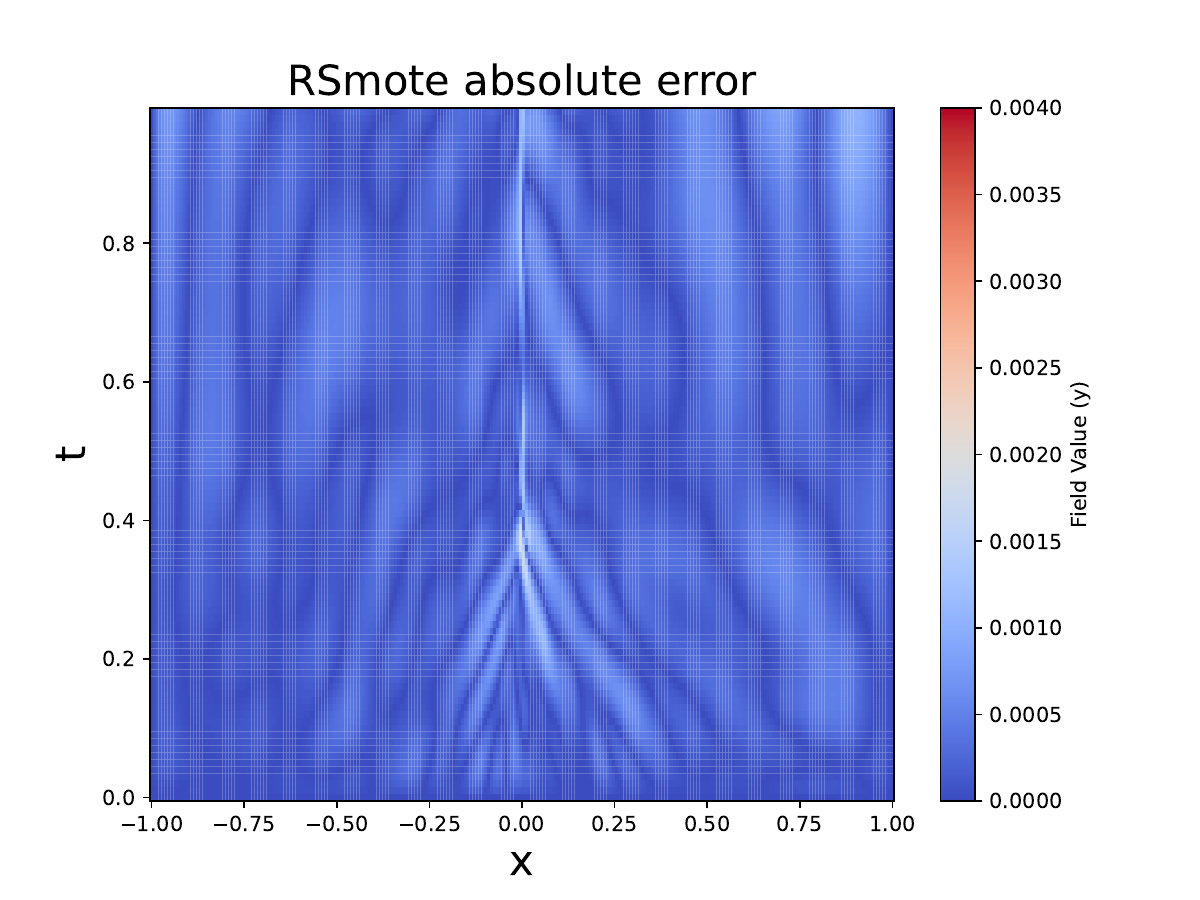}}
    \subfigure[GLF difference]{\includegraphics[width=0.32\linewidth]{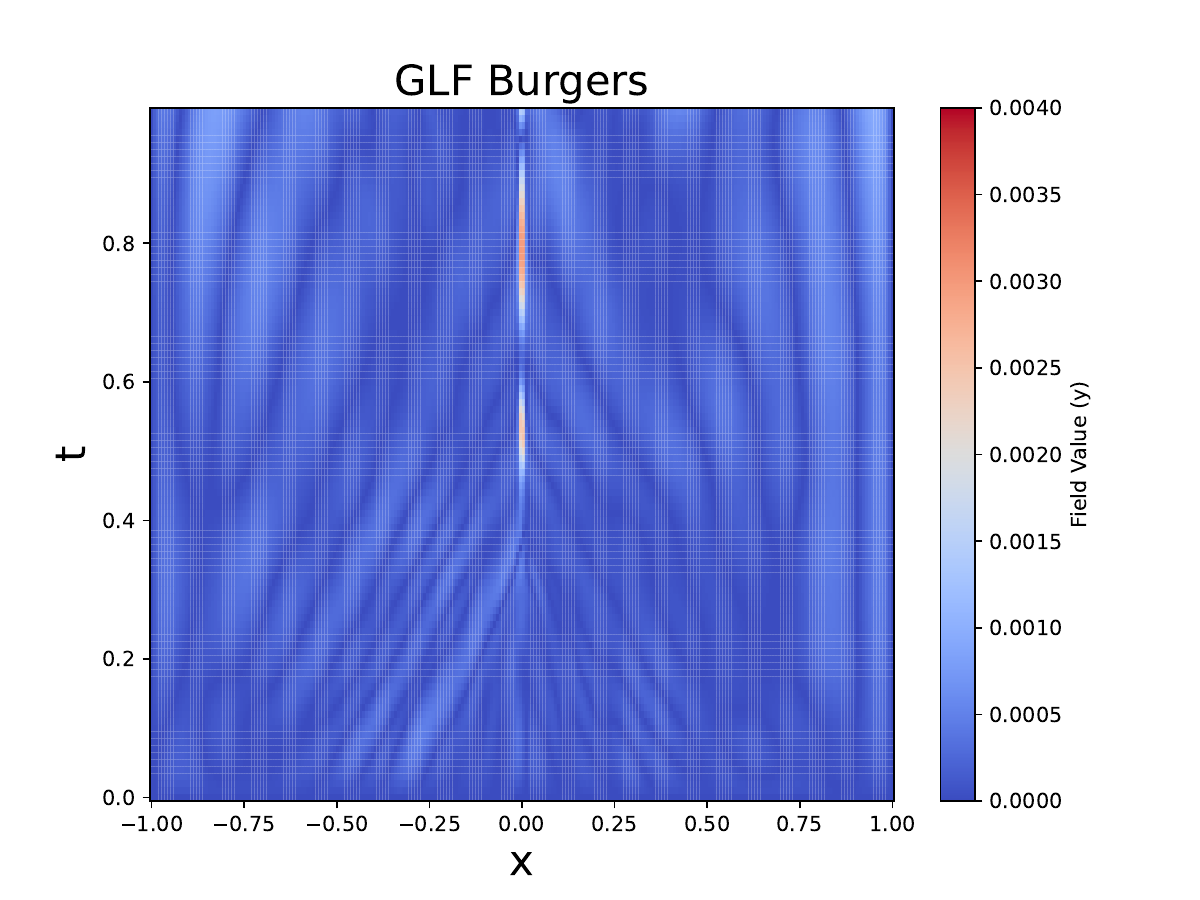}}
    \subfigure[RAD distribution]{\includegraphics[width=0.32\linewidth]{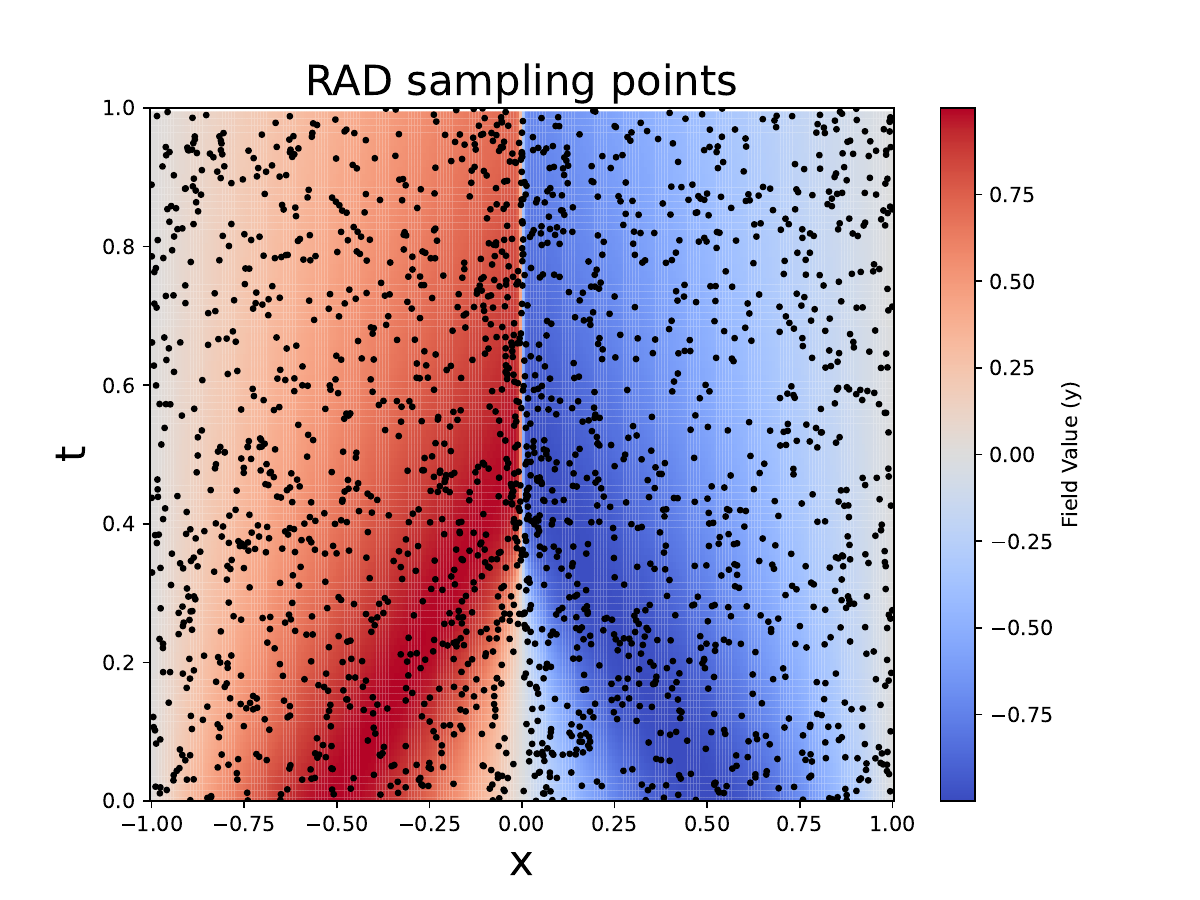}}
    \subfigure[RSmote distribution]{\includegraphics[width=0.32\linewidth]{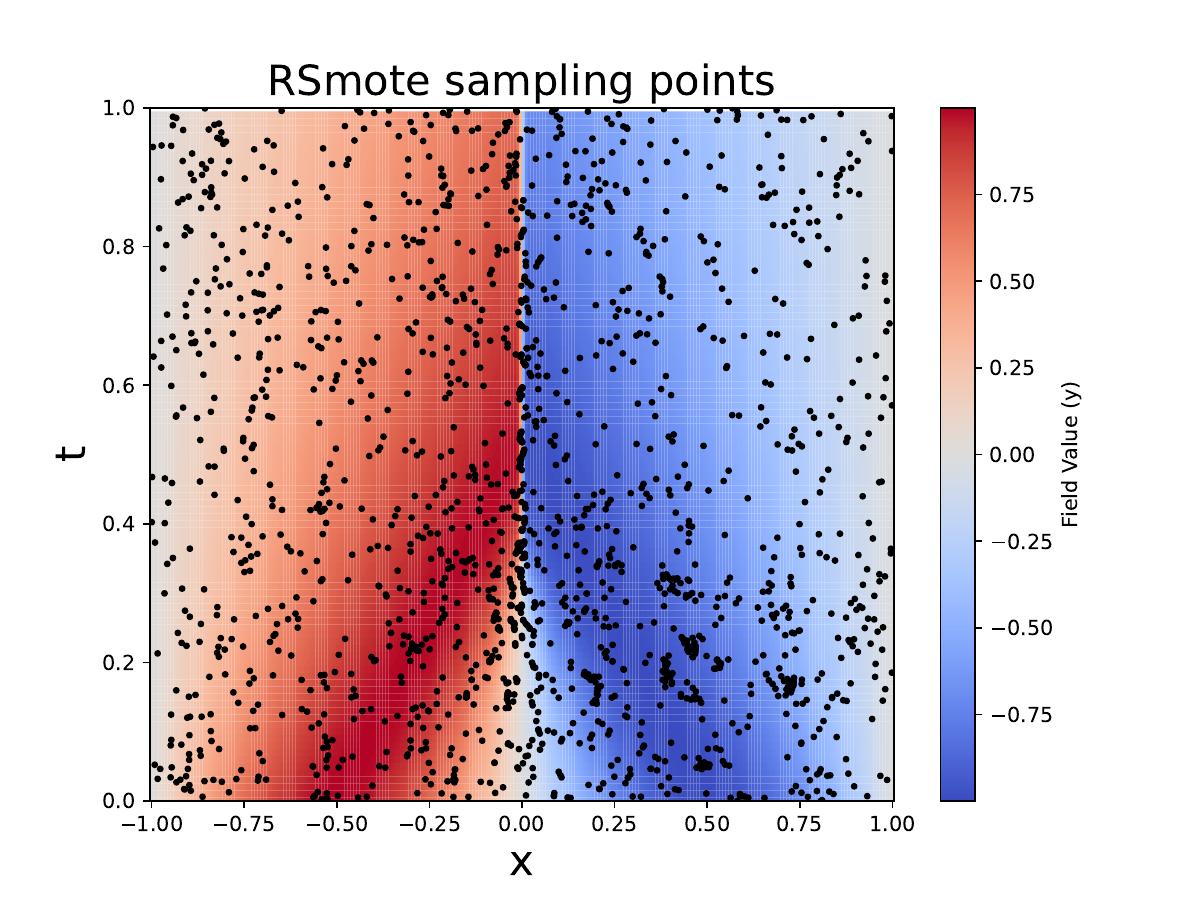}}
    \subfigure[GLF distribution]{\includegraphics[width=0.32\linewidth]{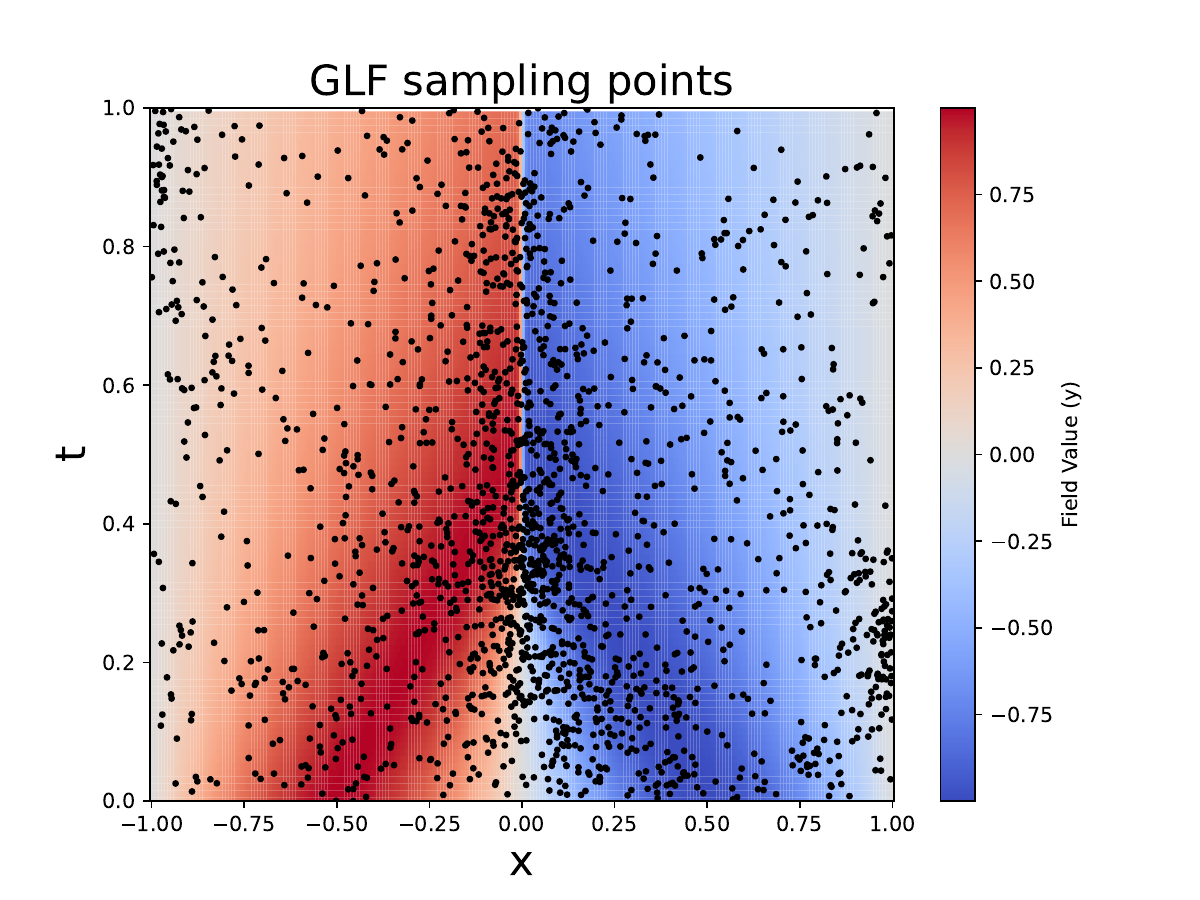}}
    \caption{Solution fields for Burgers' Equation. (a)-(c): RAD solution, RSmote solution and GLF solution; (d)-(f): Absolute differences; (g)-(i): The distribution of the final points. }
    \label{f.burgers_field}
\end{figure}

\begin{figure}[!h]
    \centering
    \subfigure[RAD solution]{\includegraphics[width=0.32\linewidth]{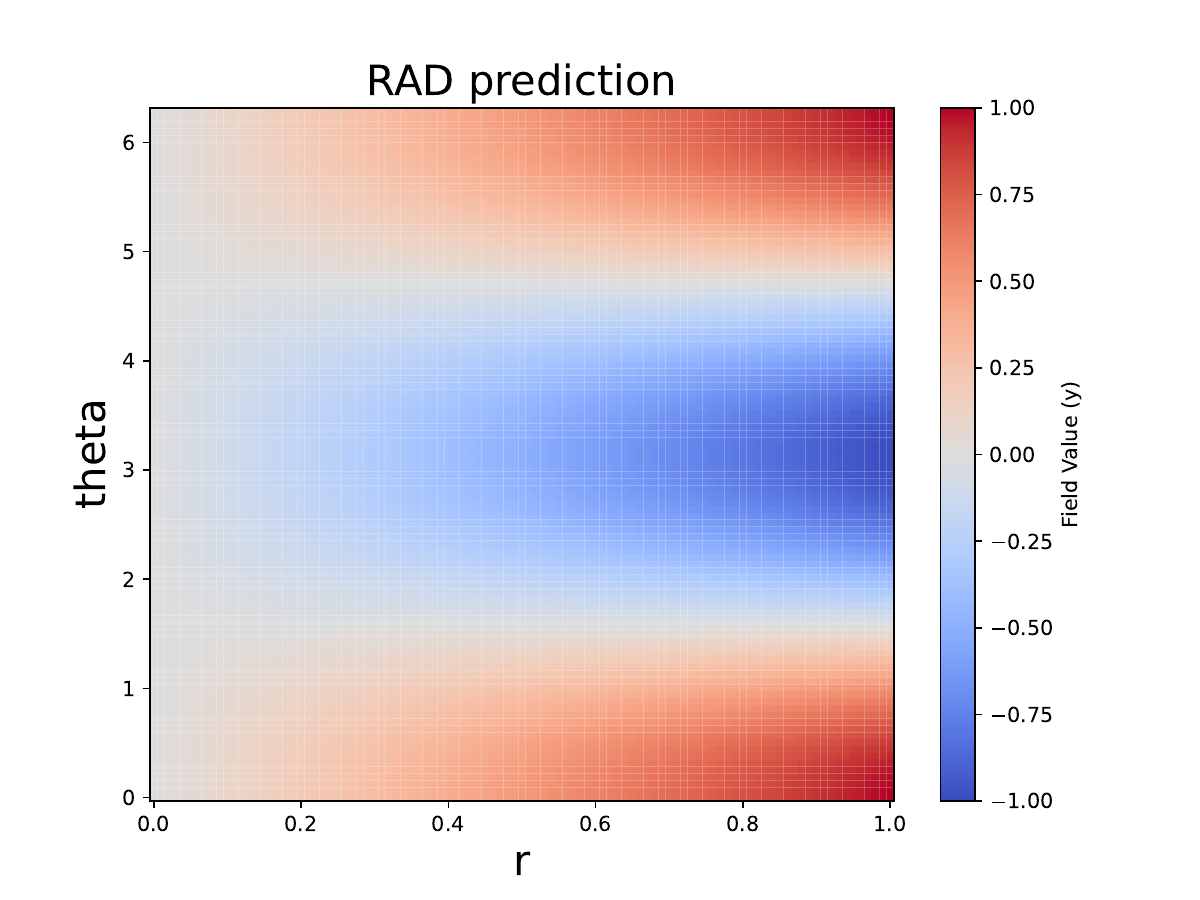}}
    \subfigure[RSmote solution]{\includegraphics[width=0.32\linewidth]{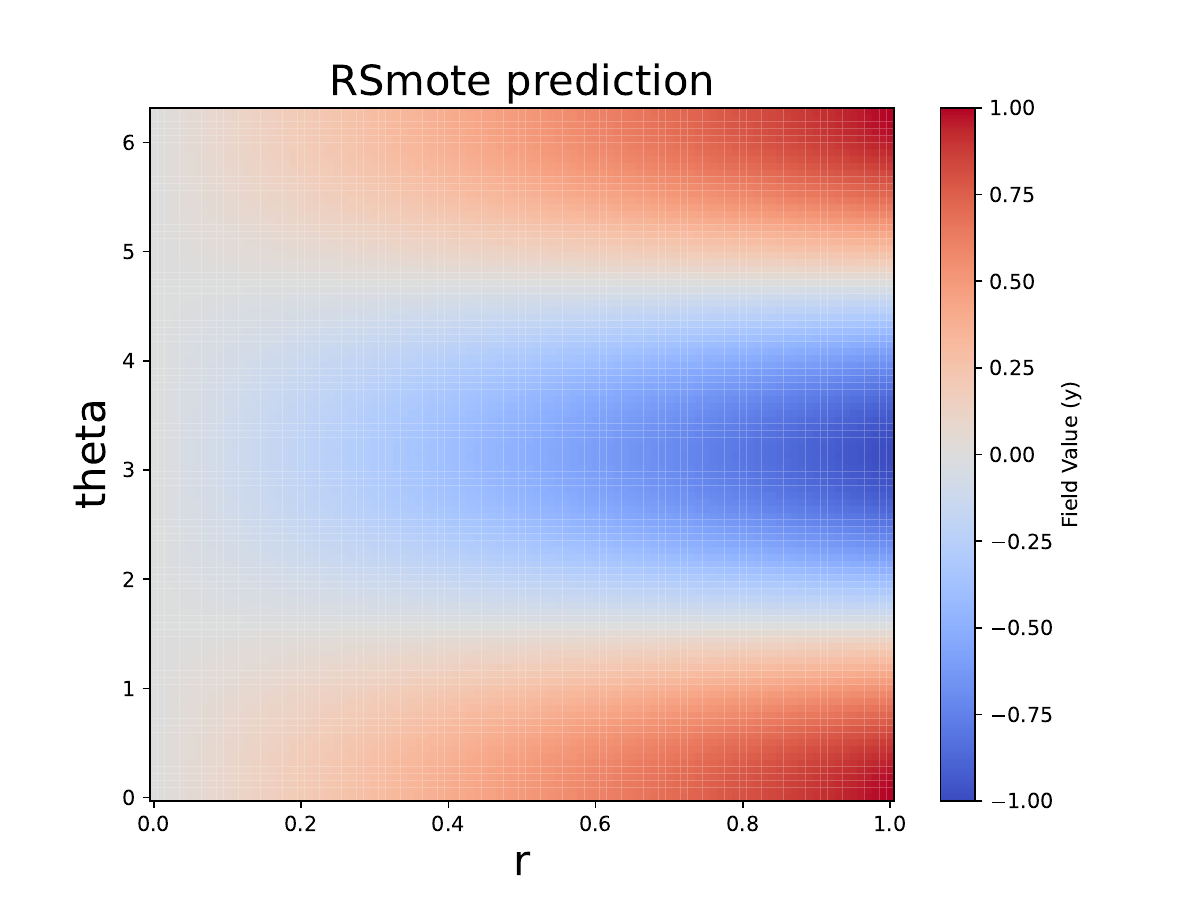}}
    \subfigure[GLF solution]{\includegraphics[width=0.32\linewidth]{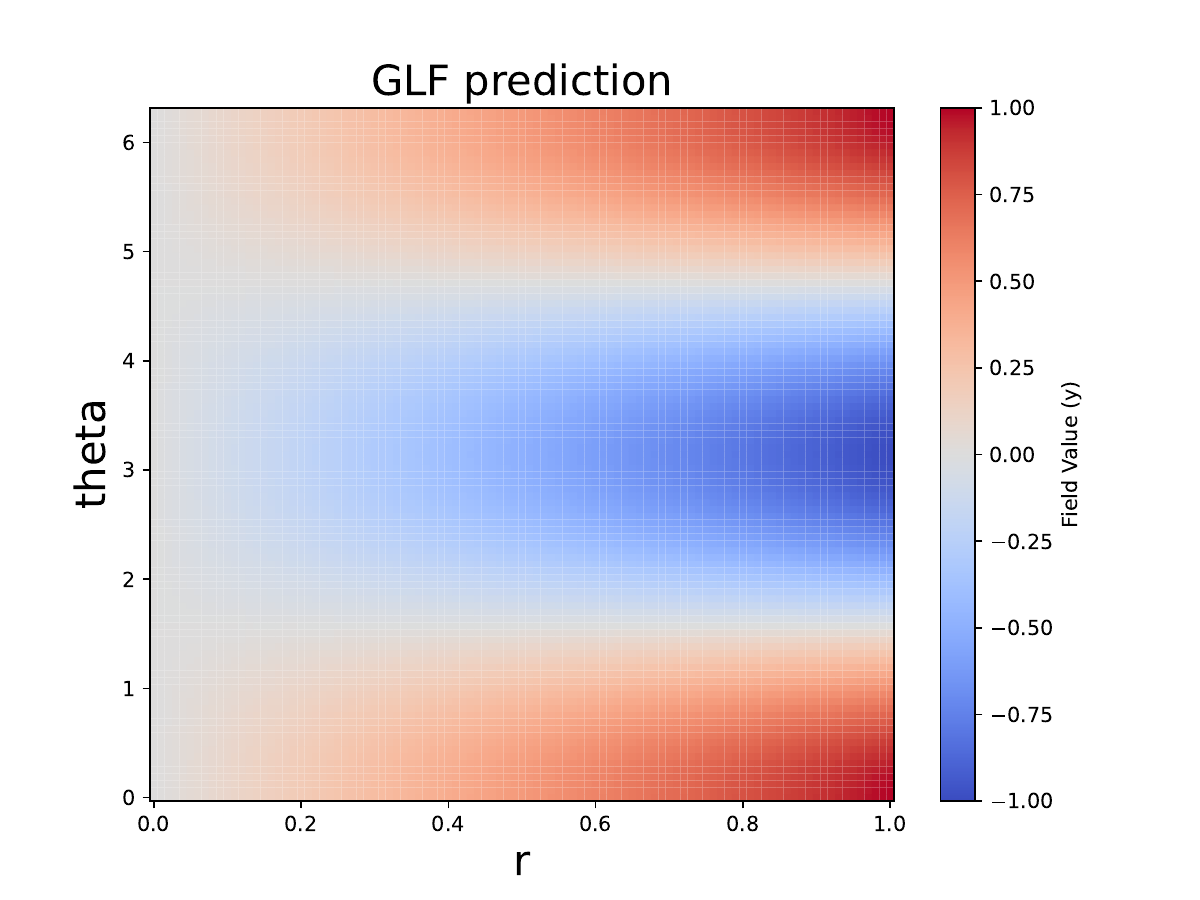}}
    \subfigure[RAD difference]{\includegraphics[width=0.32\linewidth]{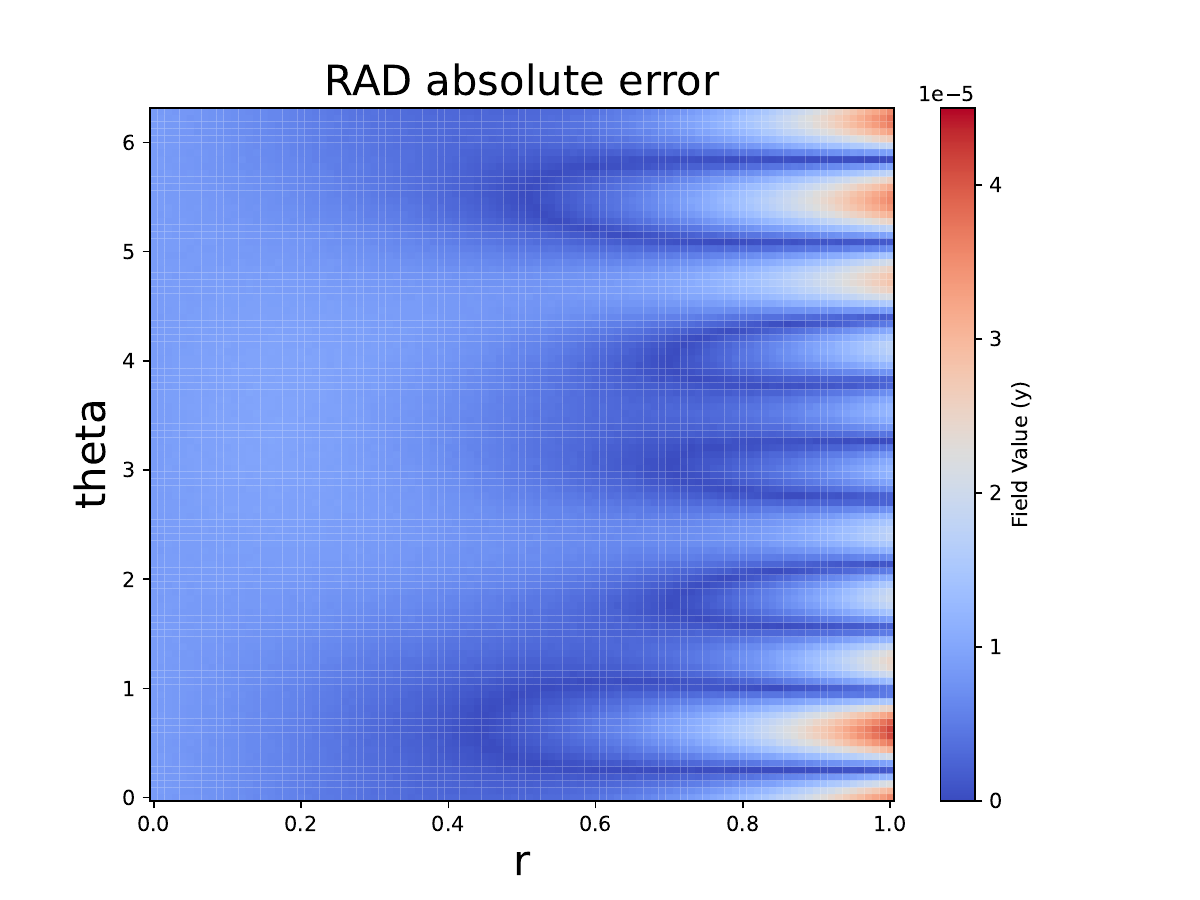}}
    \subfigure[RSmote difference]{\includegraphics[width=0.32\linewidth]{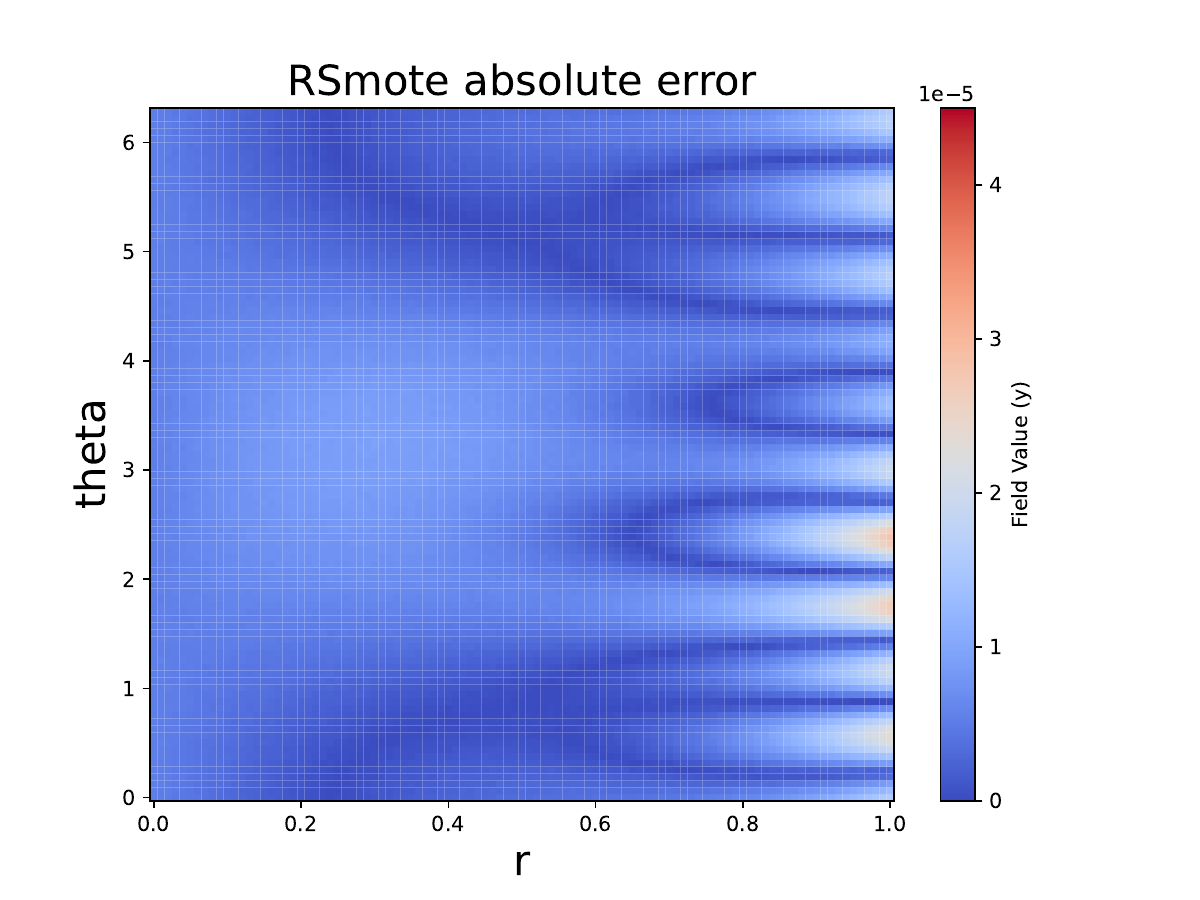}}
    \subfigure[GLF difference]{\includegraphics[width=0.32\linewidth]{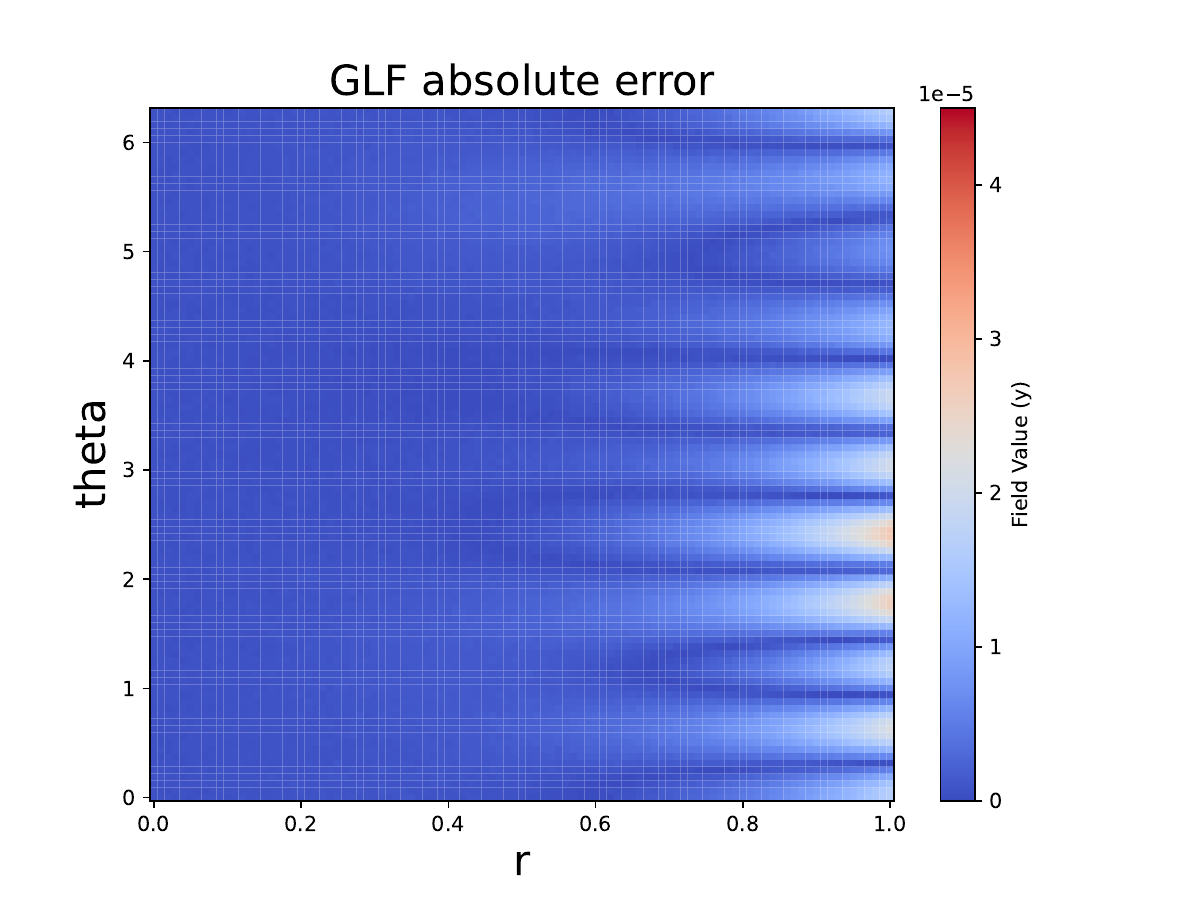}}
    \subfigure[RAD distribution]{\includegraphics[width=0.32\linewidth]{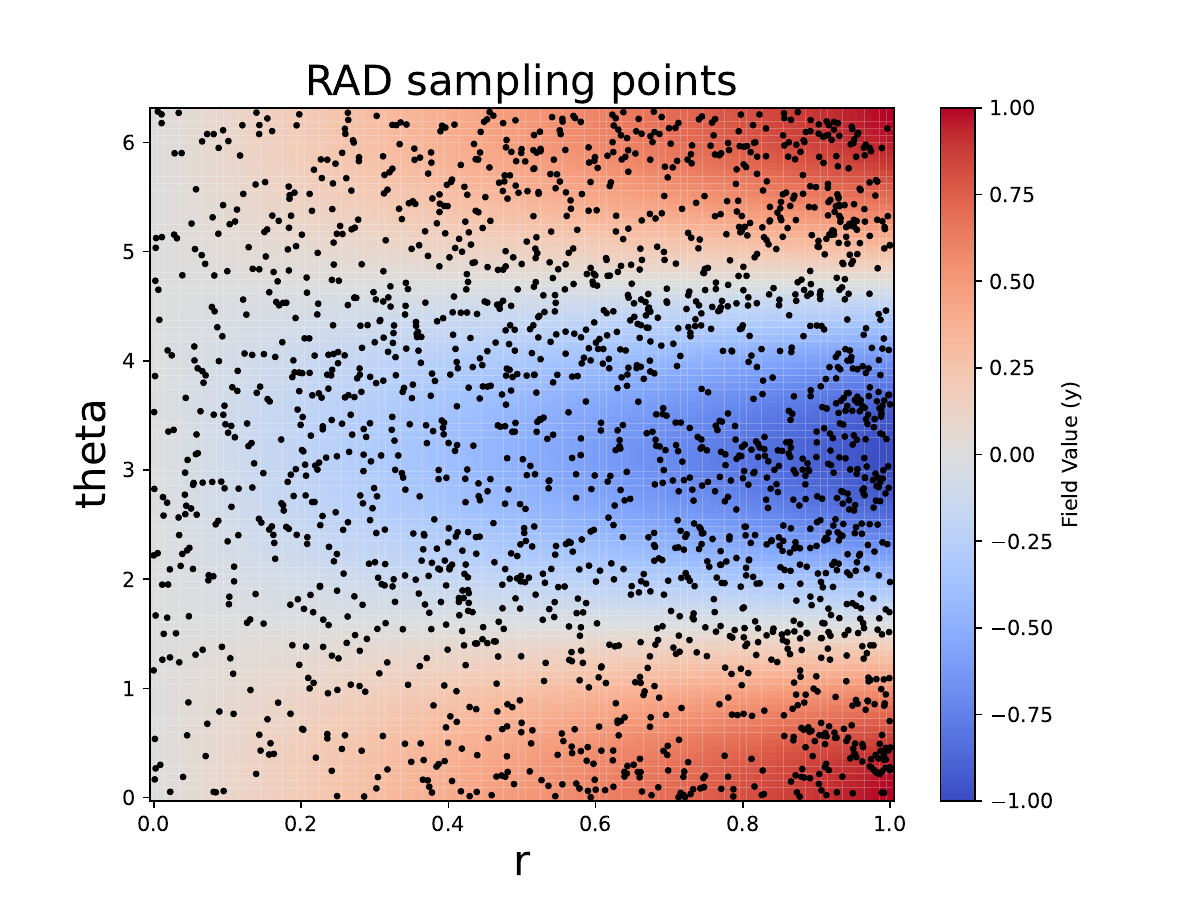}}
    \subfigure[RSmote distribution]{\includegraphics[width=0.32\linewidth]{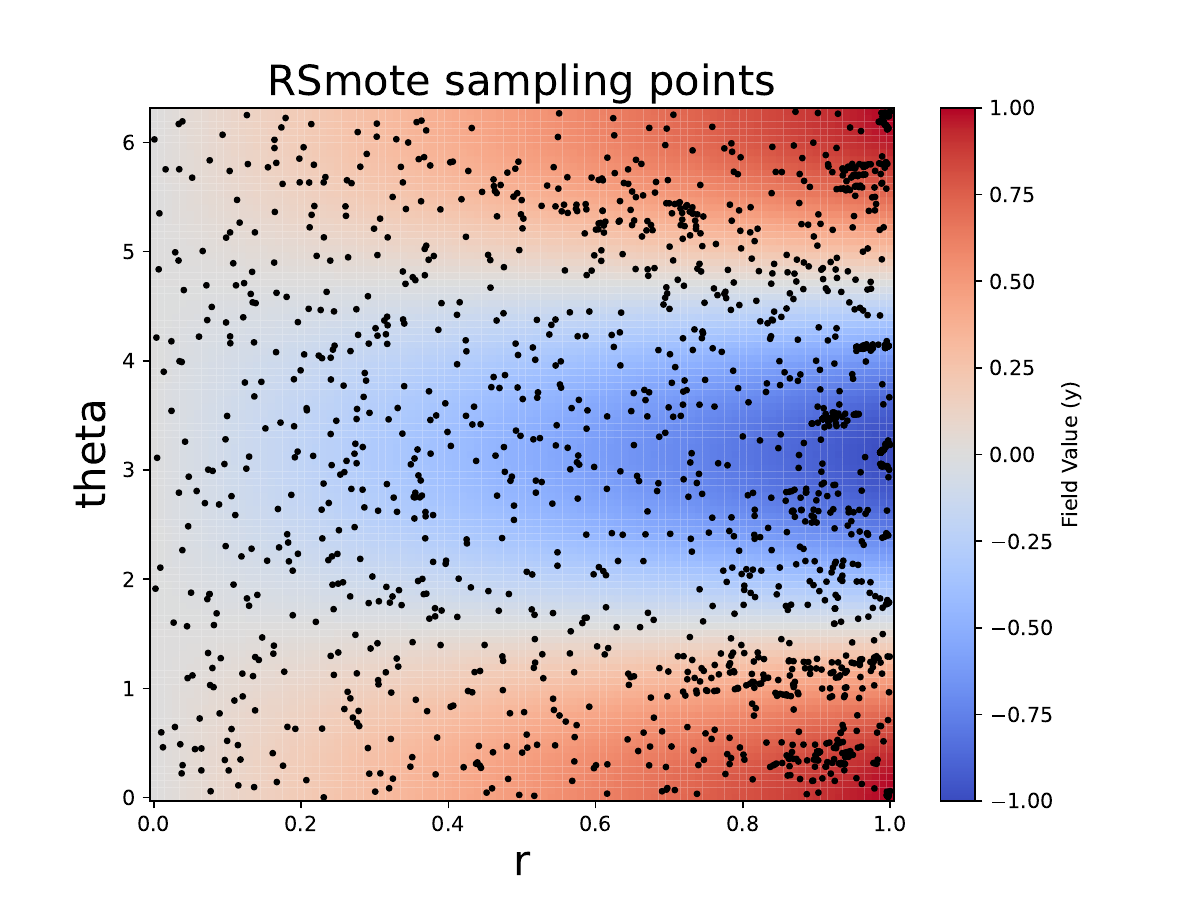}}
    \subfigure[GLF distribution]{\includegraphics[width=0.32\linewidth]{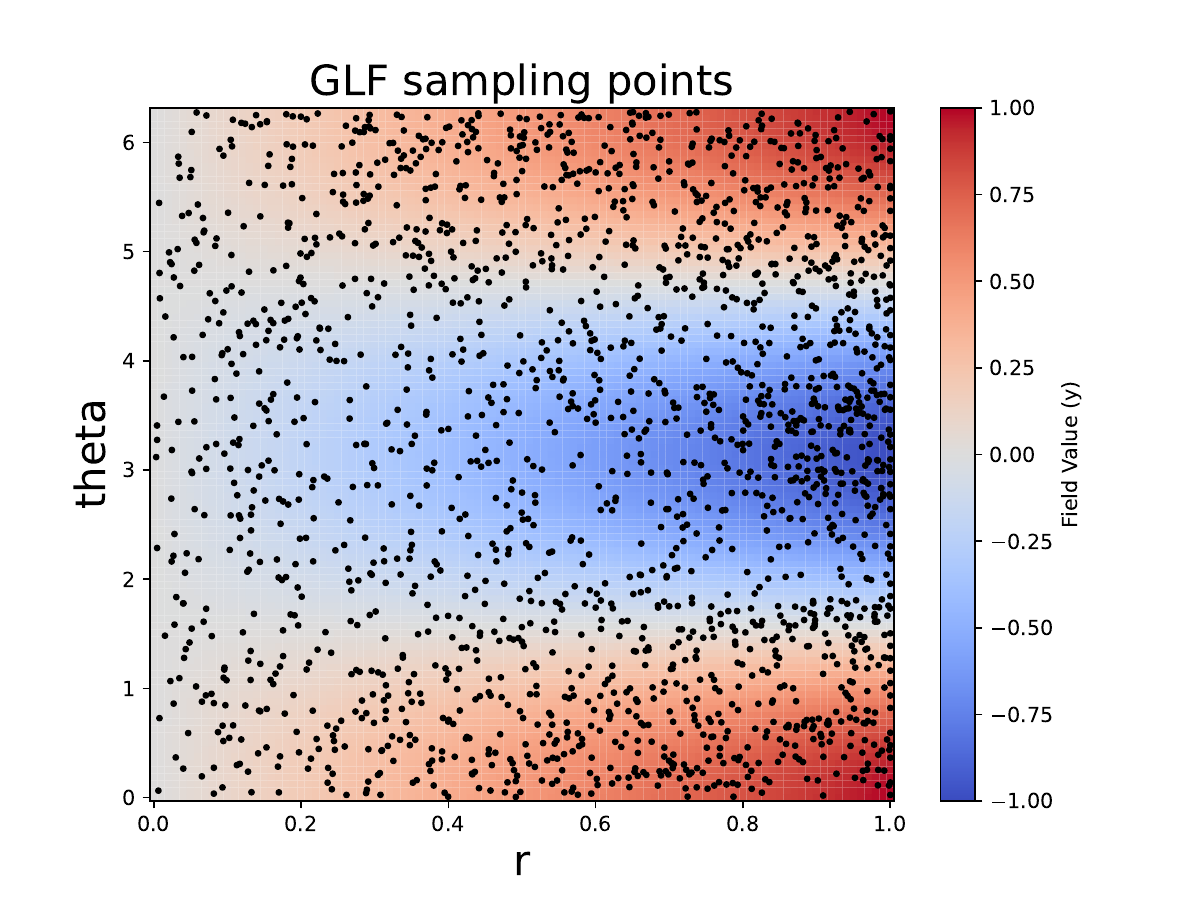}}
    \caption{Solution fields for Laplace Equation. (a)-(c): RAD solution, RSmote solution and GLF solution; (d)-(f): Absolute differences; (g)-(i): The distribution of the final points. }
    \label{f.laplace_field}
\end{figure}

To further illustrate the superiority of GLF, we provide three 2-dimensional PDE predicted solutions and absolute error maps in Fig.~\ref{f.allen_field}, Fig.~\ref{f.burgers_field}, and Fig.~\ref{f.laplace_field}. In each figure, (a)–(f) show that GLF achieves more accurate approximations across the domain, consistently yielding lower absolute errors compared to the baselines. From the distribution of the final sampling points, it is evident that the global method RAD spreads points broadly, which ensures stability but leads to redundant samples in well-learned regions. In contrast, the local method RSmote packs points in complex areas but under-samples elsewhere. GLF overcomes these limitations by adaptively concentrating samples in difficult regions while still maintaining sufficient exploration elsewhere, thereby combining the stability of global sampling with the efficiency of local refinement.

In summary, the table, loss trajectories, and field visualizations consistently confirm that GLF delivers more accurate solutions with more stable convergence and lower computational cost, combining the stability benefits of global sampling with the efficiency of local refinement.


\begin{table}[ht]
\centering
\caption{Comparison (Mean $\pm$ Std.) of GLF, RAD, and RSmote on five PDEs. The table introduces the scores and memory consumption. The lower the score, the better the performance. The \textbf{bold} indicates the best result.}
\label{tab:pde_summary}
\begin{adjustbox}{width=\textwidth}
\begin{tabular}{lcccc}
\toprule[2pt]
PDE & Metric & RAD & RSmote & GLF \\
\midrule[1.5pt] 
\multirow{2}{*}{Allen-Cahn} & Error & 0.0101 $\pm$ 0.0027 & 0.0039 $\pm$ 0.0009 & \textbf{0.0036 $\pm$ 0.0007} \\
 & Memory & 1860 & \textbf{1120} & \textbf{1120} \\
\hline
\multirow{2}{*}{Burgers}  & Error & 7.86e-4 $\pm$ 2.84e-4 & \textbf{7.83e-4 $\pm$ 2.48e-4} & 7.93e-4 $\pm$ 1.3e-4 \\
 & Memory & 1882 & \textbf{1154} & \textbf{1154} \\
\hline
\multirow{2}{*}{Laplace}  & Error & 5.83e-5 $\pm$ 2.84e-5 & 5.98e-5 $\pm$ 3.15e-5 & \textbf{4.01e-5 $\pm$ 1.09e-5} \\
 & Memory & 1558 & \textbf{1100} & \textbf{1100} \\
\hline
\multirow{2}{*}{Dispersive}  & Error & 3.35e-4 $\pm$ 1.31e-4 & 2.93e-4 $\pm$ 9.00e-5 & \textbf{2.41e-4 $\pm$ 1.05e-4} \\
 & Memory & 7164 & \textbf{1684} & \textbf{1684} \\   
\hline
\multirow{2}{*}{Reaction-Diffusion}  & Error & 7.42e-3$\pm$1.90e-4 & 5.42e-3$\pm$2.08e-4 & \textbf{4.96e-3 $\pm$ 2.95e-4} \\
 & Memory & 7286 & \textbf{1858} & \textbf{1858} \\  
\bottomrule[2pt]
\end{tabular}
\end{adjustbox}
\end{table}

\subsection{Ablation Studies}
We further investigate two design aspects of GLF:
\begin{enumerate}
\item \textbf{Neighborhood sampling size (GLF-M, i.e., GLF with more candidate points).}   Increasing the number of samples per neighborhood examines whether more local samples improves performance. For fairness, we add enough points so that the total number of candidates matches that of the global method RAD (100{,}000 points).

\item \textbf{Residual distribution strategy (GLF-D, i.e., GLF with the exact residual-based distribution).} Unlike our proposed approximation, GLF-D computes the exact residual distribution and samples points accordingly.
\end{enumerate}

Table~\ref{t.ablation} summarizes the quantitative comparison. The results show that neither modification leads to improved performance: GLF-M does not yield better accuracy, while GLF-D achieves errors nearly identical to GLF but consumes substantially more memory. 
For the Allen–Cahn and Laplace equations, all three methods achieve similar errors, yet GLF maintains the lowest memory usage. 
On Burgers’ and Reaction–Diffusion equations, GLF-D obtains slightly lower errors, but the differences from GLF are negligible relative to its additional cost. 
For the Dispersive equation, GLF delivers the best or comparable accuracy while requiring nearly half the memory of GLF-D. Furthermore, in GLF-D, the 10-dimensional third-order Dispersive equation requires a memory cost similar to that of the 20-dimensional second-order Reaction–Diffusion equation, indicating that computing higher-order derivatives is more expensive.
Overall, GLF consistently achieves the best balance between accuracy and efficiency, confirming the effectiveness of our design choices.

\begin{figure}[!h]
\centering
    \subfigure[Allen-Cahn]{\includegraphics[width=0.45\linewidth]{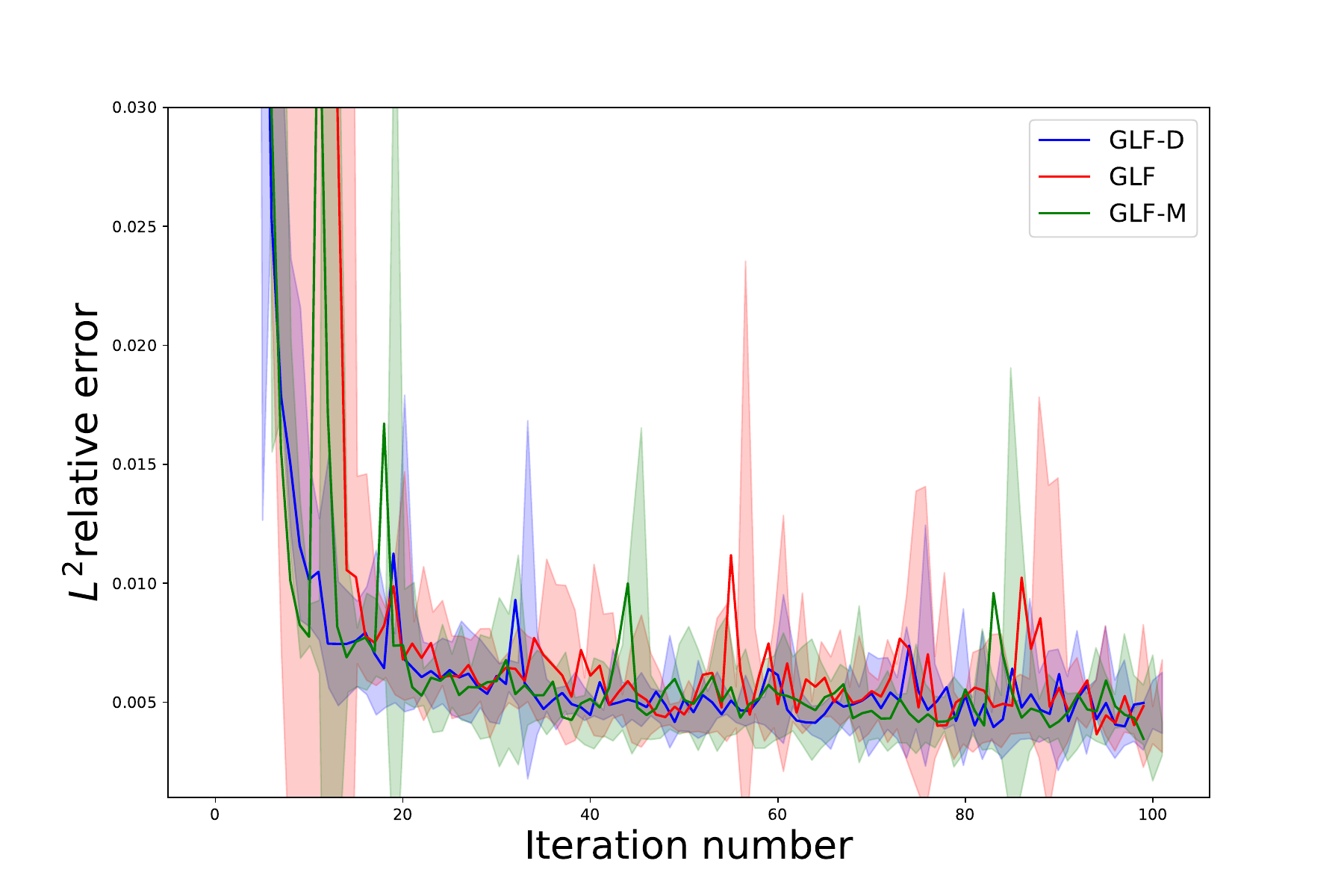}}
    \subfigure[Burgers]{\includegraphics[width=0.45\linewidth]{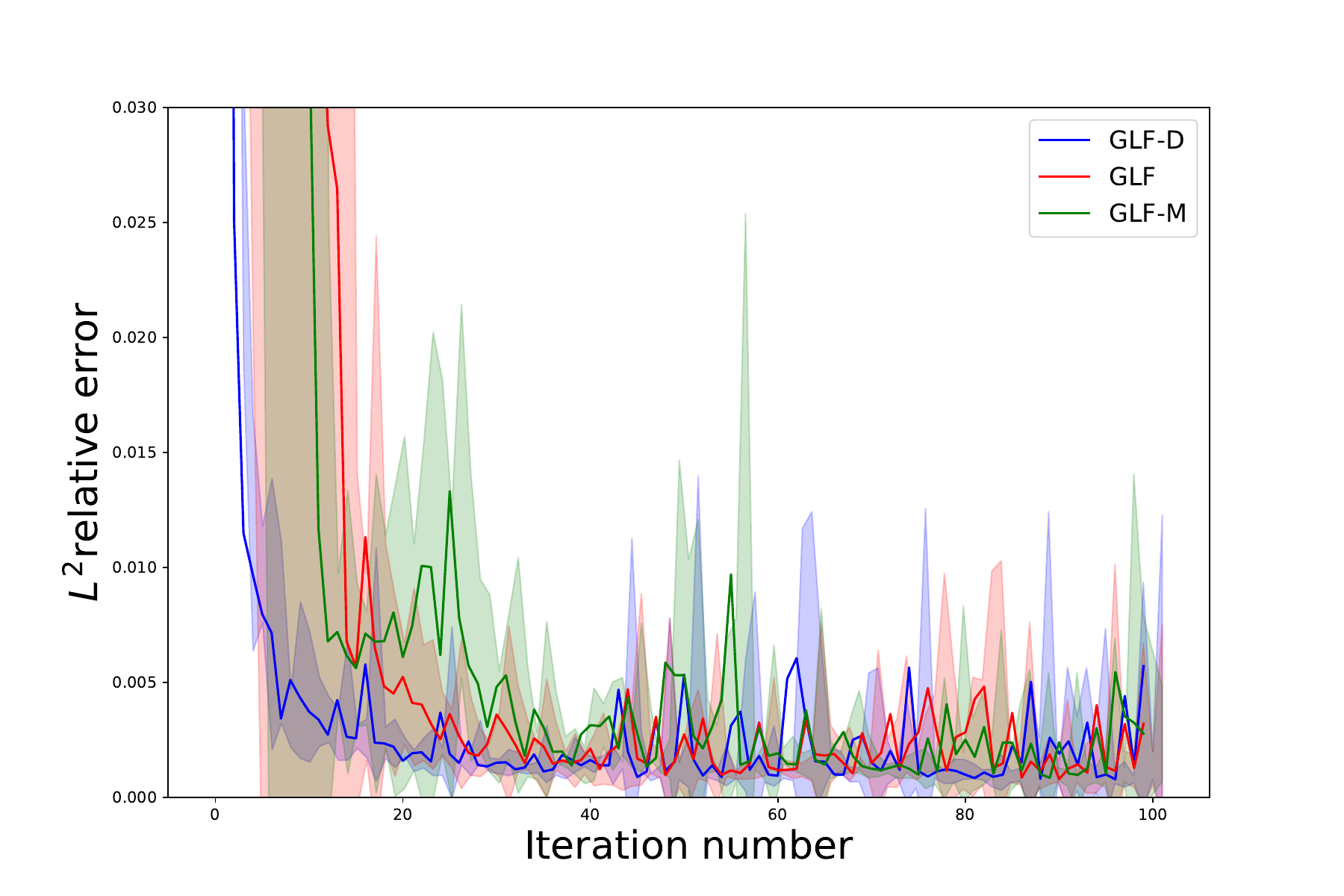}}
    \subfigure[Laplace]{\includegraphics[width=0.45\linewidth]{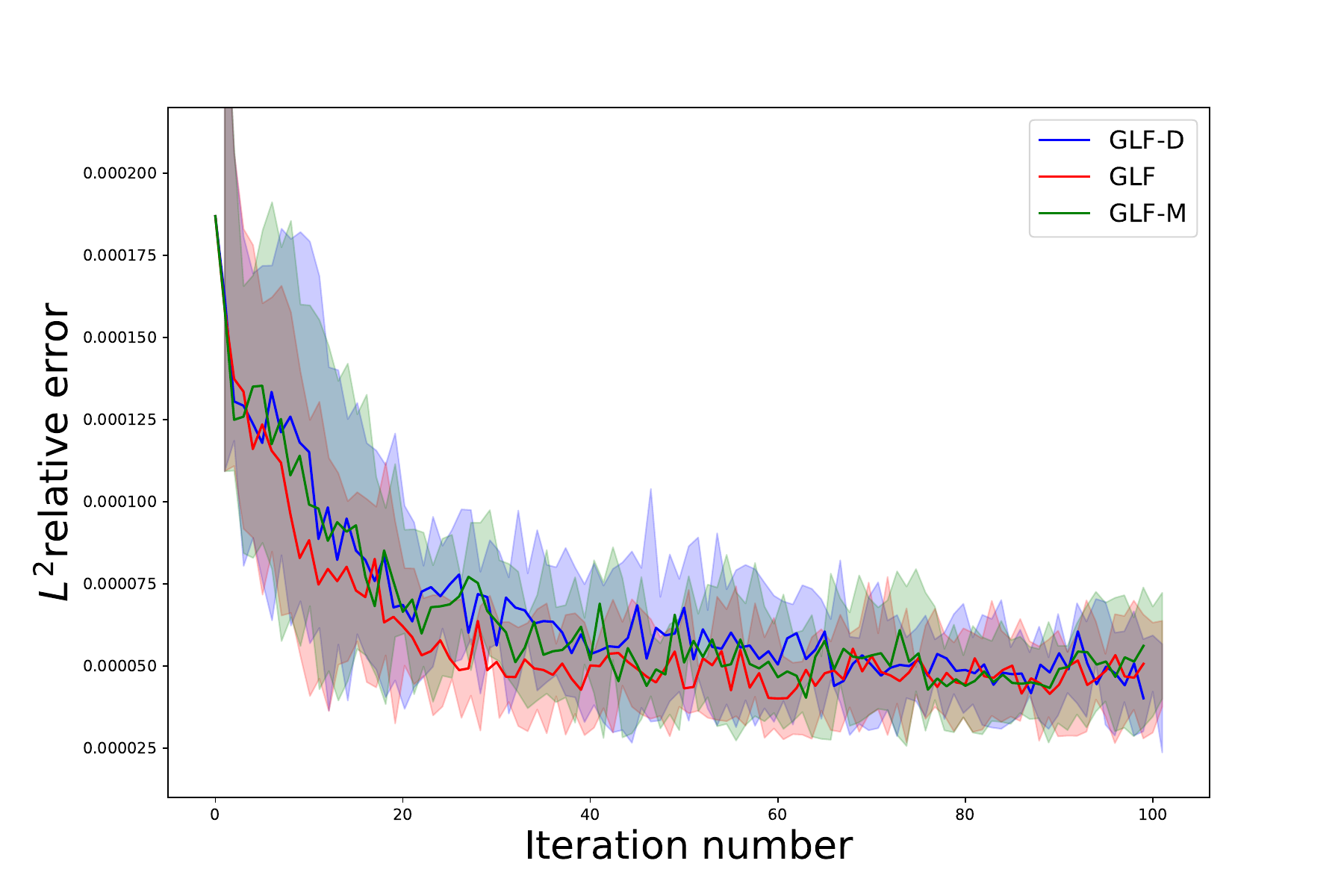}}
    \subfigure[Dispersive]{\includegraphics[width=0.45\linewidth]{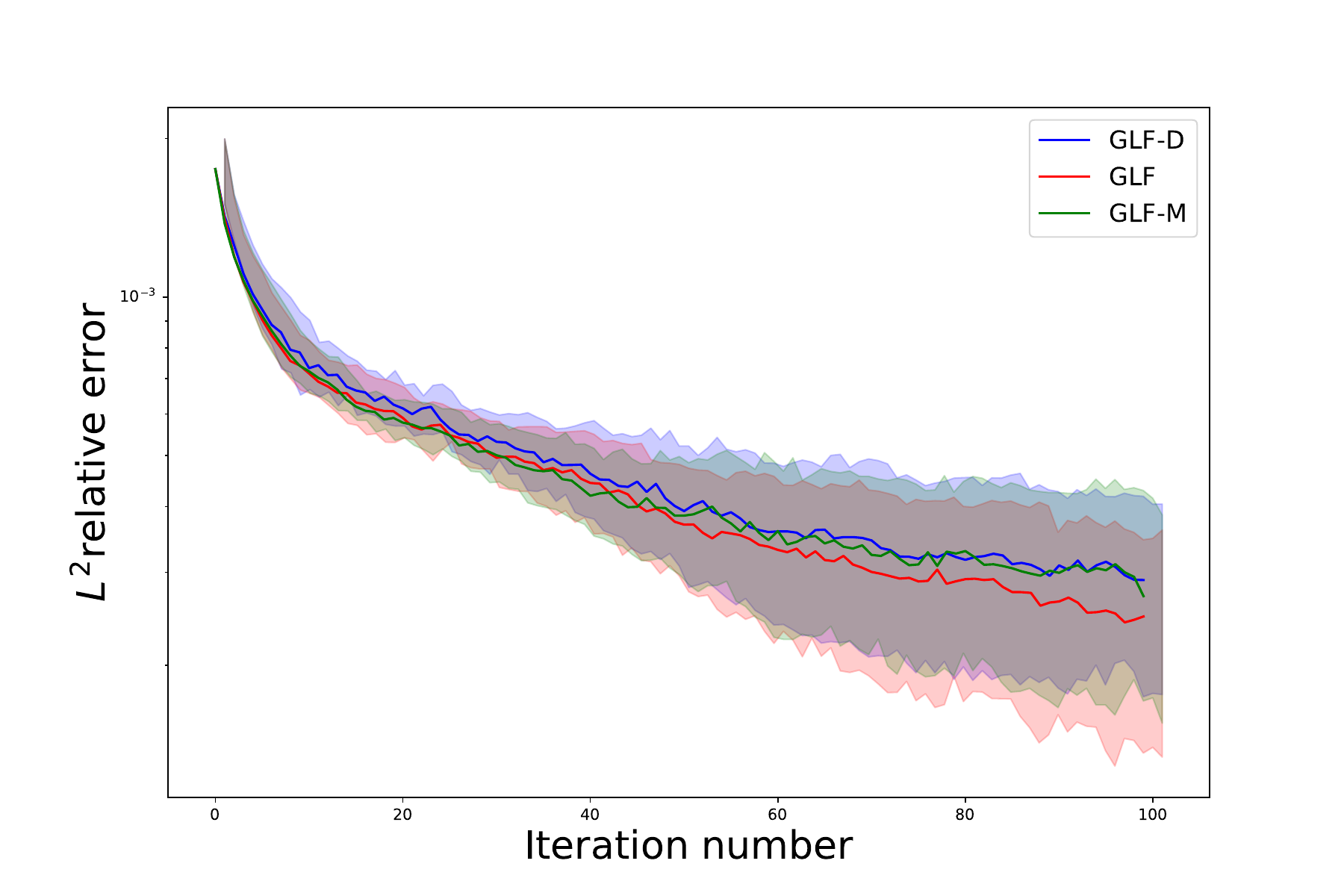}}
    \subfigure[Reaction-Diffusion]{\includegraphics[width=0.45\linewidth]{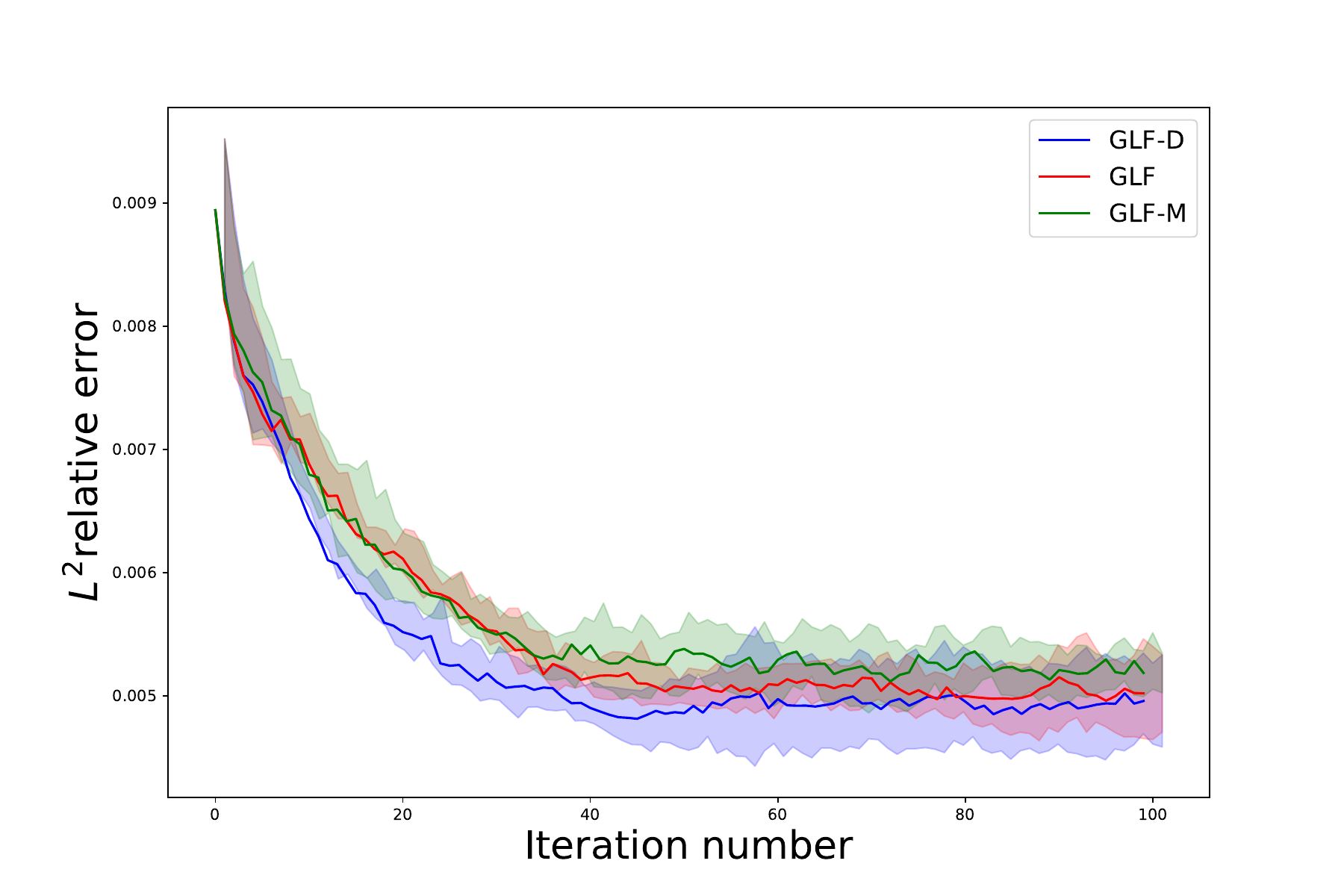}}
\caption{Training loss curves for different PDEs. Red line: Mean values of GLF method; Blue line: Mean values of GLF-D method; Green line: Mean values of GLF-M method. The shaded areas represent the corresponding standard deviations.}
\label{f.ablation}
\end{figure}

Fig.~\ref{f.ablation} further illustrates the training losses on five PDEs. 
The results confirm that increasing the neighborhood sample size does not improve performance and may even degrade it, while the approximate residual distribution used in GLF achieves accuracy comparable to the exact PMF at a fraction of the memory cost. 
Together with the quantitative results, these findings validate that GLF effectively integrates global exploration and local refinement while avoiding unnecessary computational overhead.

\begin{table}[!h]
\centering
\caption{Comparison (Mean $\pm$ Std.) of GLF, GLF-M, and GLF-D on five PDEs. The table shows the scores and memory consumption. The lower the score, the better the performance. The \textbf{bold} indicates the best result.}
\label{t.ablation}
\begin{adjustbox}{width=\textwidth}
\begin{tabular}{lcccc}
\toprule[2pt]
PDE & Metric & GLF-D & GLF-M & GLF \\
\midrule[1.5pt] 
\multirow{2}{*}{Allen-Cahn} & Error & 0.0039 $\pm$ 0.0009 & 0.0037 $\pm$ 0.0007 & \textbf{0.0036 $\pm$ 0.0007} \\
 & Memory & 1166 & \textbf{1120} & \textbf{1120} \\
\hline
\multirow{2}{*}{Burgers}  & Error & \textbf{7.86e-4 $\pm$ 1.58e-4} & 8.51e-4 $\pm$ 2.48e-4 & 7.93e-4 $\pm$ 1.3e-4 \\
 & Memory & 1198 & \textbf{1154} & \textbf{1154} \\
\hline
\multirow{2}{*}{Laplace}  & Error & 4.02e-5 $\pm$ 1.65e-5 & 4.03e-5 $\pm$ 1.25e-5 & \textbf{4.01e-5 $\pm$ 1.09e-5} \\
 & Memory & 1122 & \textbf{1100} & \textbf{1100} \\
\hline
\multirow{2}{*}{Dispersive}  & Error & 2.80e-4 $\pm$ 3.87e-5 & 2.70e-4 $\pm$ 1.15e-4 & \textbf{2.41e-4 $\pm$ 1.05e-4} \\
 & Memory & 3244 & \textbf{1684} & \textbf{1684} \\   
\hline
\multirow{2}{*}{Reaction-Diffusion}  & Error & \textbf{4.81e-3$\pm$2.67e-4} & 5.12e-3$\pm$2.47e-4 & 4.96e-4 $\pm$ 2.95e-4 \\
 & Memory & 3478 & \textbf{1858} & \textbf{1858} \\  
\bottomrule[2pt]
\end{tabular}
\end{adjustbox}
\end{table}

\section{Conclusions}
\label{s:conclu}

This paper introduced a global–local fusion sampling strategy that addresses the trade-off between global exploration and local refinement in PINNs. The main contributions are threefold. First, we propose a residual-adaptive neighborhood sampling method that allocates more points to high-residual regions while maintaining broader exploration elsewhere. Second, we replace the costly global residual-based distribution with a lightweight approximation, preserving the stable convergence characteristic of global sampling while significantly reducing computational cost. Finally, comprehensive benchmark experiments demonstrate that GLF inherits the stability of global methods and the efficiency of local refinement, achieving higher accuracy with lower memory consumption.
Ablation studies further validate that enlarging neighborhood sample sizes or computing exact residual distributions does not yield meaningful benefits, confirming that GLF’s efficiency-centric choices are the right ones. In sum, GLF is more accurate, more stable, and more efficient than existing global or local samplers. Looking ahead, the principles of global–local fusion may extend beyond PDEs to other deep learning-based methods where sampling efficiency is a bottleneck.





\bibliographystyle{elsarticle-num} 
\bibliography{references}

\end{document}